%% file: main.tex
\documentclass[10pt,twocolumn,letterpaper]{article}

\usepackage{cvpr}              
\usepackage{xcolor} 
\definecolor{stageblue}{RGB}{68,114,196}
\definecolor{pptGreen6}{RGB}{112,173,71}
\usepackage{amssymb}
\usepackage{adjustbox}

\usepackage[table]{xcolor}
\usepackage{pifont}
\usepackage{subcaption}
\usepackage[table]{xcolor}

\input{preamble}
\definecolor{cvprblue}{rgb}{0.21,0.49,0.74}
\usepackage[pagebackref,breaklinks,colorlinks,allcolors=cvprblue]{hyperref}
\usepackage{multirow}
\usepackage{pifont} 
\usepackage{colortbl}
\usepackage{xcolor}
\usepackage{tabularx}

\definecolor{lightblue}{RGB}{230,242,255}

\title{Semantics Lead the Way: Harmonizing Semantic and Texture Modeling \\ with Asynchronous Latent Diffusion}

\author{
Yueming Pan\textsuperscript{1,2 * ‡},
Ruoyu Feng\textsuperscript{3 ‡},
Qi Dai\textsuperscript{2},
Yuqi Wang\textsuperscript{3},\\
Wenfeng Lin\textsuperscript{3},
Mingyu Guo\textsuperscript{3},
Chong Luo\textsuperscript{2 †},
Nanning Zheng\textsuperscript{1 †}\\[3pt]
\textsuperscript{1}IAIR, Xi’an Jiaotong University \quad
\textsuperscript{2}Microsoft Research Asia
\quad
\textsuperscript{3}ByteDance 
}

\begin{document}

\twocolumn[{
\maketitle
\vspace{-3.5em}
\renewcommand\twocolumn[1][]{#1}

\begin{center}
    \centering
    \includegraphics[width=1\textwidth]{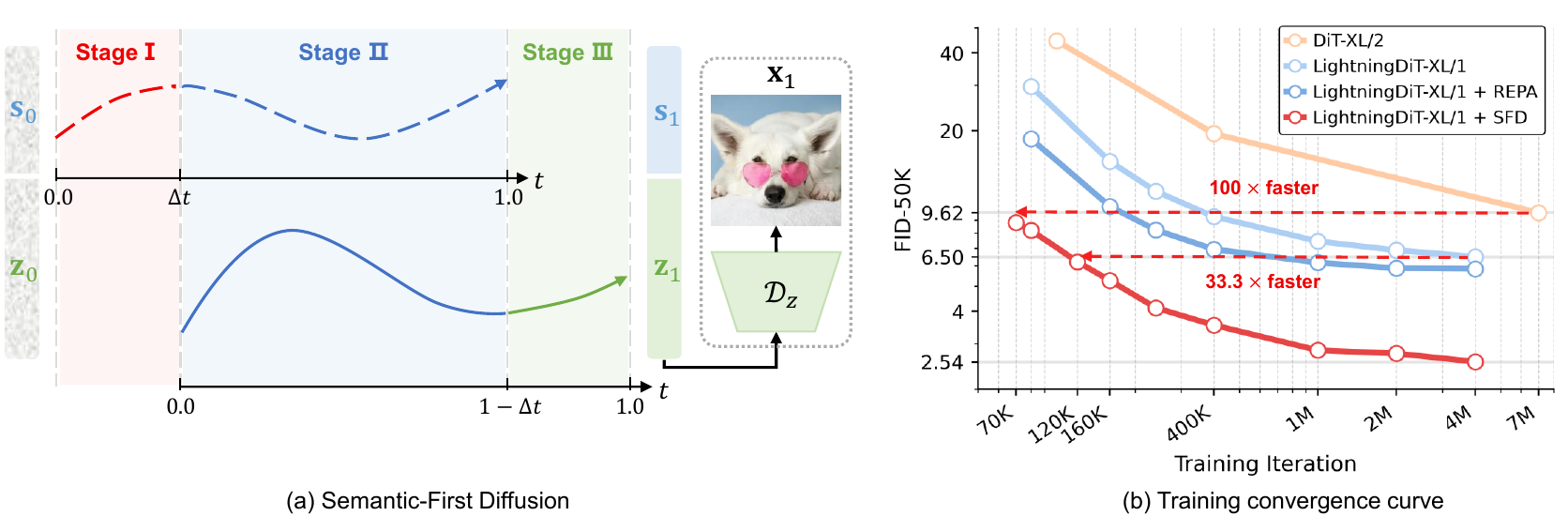}
    \vspace{-0.6cm}
    \captionsetup{hypcap=false}
    \captionof{figure}{
    \textbf{(a) Overview of Semantic-First Diffusion (SFD).} 
    Semantics (dashed curve) and textures (solid curve) follow asynchronous denoising trajectories.
    SFD operates in three phases: \textcolor{red}{Stage $\mathrm{I}$ -- Semantic initialization}, where semantic latents denoise first;
    \textcolor{stageblue}{Stage $\mathrm{II}$ -- Asynchronous generation}, where semantics and textures denoise jointly but asynchronously, with semantics ahead of textures; 
    \textcolor{green!60!black}{Stage $\mathrm{III}$ -- Texture completion}, where only textures continue refining.
    After denoising, the generated semantic latent $\mathbf{s}_1$ is discarded, and the final image is decoded solely from the texture latent $\mathbf{z}_1$.
    \textbf{(b) Training convergence on ImageNet 256$\times$256 without guidance.} 
    SFD achieves substantially faster convergence than DiT-XL/2 and LightningDiT-XL/1 by approximately \textbf{100}$\times$ and \textbf{33.3}$\times$, respectively.
    }
    \label{fig:teaser}
\end{center}

}
]

\input{sec/0_abstract}

\input{sec/1_intro}

\input{sec/2_relatedwork}
\input{sec/3_method}
\input{sec/4_experiments}

\input{sec/5_conclusion}
\section*{Acknowledgments}
Yueming Pan and Nanning Zheng were supported in part by the NSFC under Grant No.~62088102.

{
    \small
    \bibliographystyle{ieeenat_fullname}
    \bibliography{main}
}

\input{sec/X_suppl}

\end{document}

%% file: sec/0_abstract.tex
\begingroup
\renewcommand\thefootnote{\fnsymbol{footnote}}
\setcounter{footnote}{0}
\footnotetext[1]{This work was performed during Yueming Pan's internship at MSRA.}
\footnotetext[3]{Equal contribution.}
\footnotetext[2]{Corresponding author.}
\endgroup

\begin{abstract}

Latent Diffusion Models (LDMs) inherently follow a coarse-to-fine generation process, where high-level semantic structure is generated slightly earlier than fine-grained texture.
This indicates the preceding semantics potentially benefit the texture generation by providing a semantic anchor.
Recent advances have integrated semantic priors from pretrained visual encoders to further enhance LDMs, yet they still denoise semantic and
VAE-encoded texture synchronously, neglecting such ordering.
Observing these, we propose \textbf{\textit{Semantic-First Diffusion (SFD)}}, a latent diffusion paradigm that explicitly prioritizes semantic formation.
SFD first constructs composite latents by combining the compact semantic latent, which is extracted from pretrained visual encoder via a dedicated \textbf{\textit{Semantic VAE}}, with the texture latent.
The core of SFD is to denoise the semantic and texture latents asynchronously using separate noise schedules: semantics precede textures by a temporal offset, providing clearer high-level guidance for texture refinement and enabling natural coarse-to-fine generation.
On ImageNet 256$\times$256 with guidance, SFD achieves \textbf{FID 1.06} (LightningDiT-XL) and \textbf{FID 1.04} (1.0B LightningDiT-XXL), while achieving up to \textbf{100}$\times$ faster convergence than original DiT without guidance.
SFD also improves existing methods like ReDi and VA-VAE, demonstrating the effectiveness of asynchronous, semantics-led modeling. Project page and code: \url{https://yuemingpan.github.io/SFD.github.io/}.

\end{abstract}

%% file: sec/1_intro.tex
\vspace{-4mm}
\section{Introduction}
\label{sec:intro}

Latent Diffusion Models (LDMs)~\cite{ldm} have emerged as the leading approach for modeling visual signals, demonstrating remarkable performance in high-quality image synthesis~\cite{ldm,dit,sit}. LDMs comprise two key components: a Variational Autoencoder (VAE)~\cite{vae} that compresses high-dimensional visual signals into a compact latent space, and a diffusion model that learns the distribution of this latent space. However, this design presents an inherent challenge. The VAE, optimized for pixel-level reconstruction, predominantly captures low-level texture features in its latent representation. 
Consequently, the diffusion model faces a conflicting objective: it must simultaneously capture high-level semantic understanding while preserving low-level textural details, which leads to slow convergence and suboptimal generation quality.

To overcome these challenges, recent studies enhance LDMs with discriminative semantic priors from pretrained visual encoders, enabling faster convergence and improved generation quality. These approaches typically achieve this by explicitly aligning semantic representations with the VAE latent space~\cite{vavae} or diffusion intermediate features~\cite{REPA, repa-e}, or by jointly modeling concatenated semantic and texture representations within the diffusion process~\cite{reg, redi}.
All these methods share a similar paradigm, i.e., all latent information, including high-level semantics and low-level textures, is denoised synchronously at the same noise level throughout the diffusion process.
However, such design deviates from the inherent nature of diffusion model, which follows a coarse-to-fine mechanism that progressively generates low-frequency structure before high-frequency textures~\cite{dctdiff, Diffusability,ddt,rissanen2023generative}.
Inspired by this natural property, we argue that discriminative semantics, which capture structural and high-level information, should not only be embedded in the latent space as part of the denoising target, but should also actively lead the generation process by evolving earlier than textures. This philosophy is akin to the principle that one should first draw a blueprint before engaging in fine decoration, rather than attempting to simultaneously define structure and detail from chaos.

In this paper, we propose to explicitly intervene in the order of information formation during generation, where discriminative semantics are synthesized first and serve as priors to guide texture generation.
However, a hard sequential generation scheme would exhibit training-inference mismatch similar to exposure bias in teacher forcing~\cite{ranzato2015sequenceICLR2016}: the model is trained with ground-truth semantic conditions but must generate based on its own imperfect predictions at inference, leading to performance degradation.
To address this, we introduce asynchronous diffusion to harmonize the joint modeling of semantics and textures: semantics evolve ahead to guide texture synthesis, while both denoise simultaneously at different noise levels.

Motivated by these insights, we propose Semantics-First Diffusion (SFD), a new paradigm for LDMs that consists of two key components. First, an explicitly constructed \textit{composite latent space} that combines discriminative semantics and low-level textures. Second, \textit{asynchronous diffusion} guided by cleaner semantic information. Specifically, for the first component, building upon a texture VAE (e.g., SD-VAE), we introduce a dedicated Semantic VAE (SemVAE) that compresses high-dimensional semantic representations from pretrained visual encoders into a compact latent space, which is then concatenated with the texture latents. For the second component, as illustrated in Figure~\ref{fig:teaser} (a), a three-stage asynchronous denoising process is proposed. In the first stage, only the semantic latents are denoised, allowing the model to establish a coarse global layout initially. In the second stage, semantics and textures are denoised jointly but at different noise levels. Since semantic features evolve ahead, they provide stronger global guidance for texture refinement. In the third stage, with semantics fully denoised, only textures continue refining details. Finally, the output image is decoded solely from the texture latent.

Our contributions are summarized as follows:
\begin{itemize}
    \item We design a composite latent space composed of semantic latents from a dedicated Semantic VAE and texture latents from SD-VAE, where the Semantic VAE compresses high-level features from pretrained visual encoders into compact representations while largely preserving semantic integrity and spatial layout.
    \item We propose the semantic-first asynchronous diffusion mechanism, which employs a three-stage denoising schedule where semantics evolve earlier and subsequently guide texture generation.
    \item SFD achieves state-of-the-art FID score of 1.04 on ImageNet 256$\times$256, while demonstrating 100$\times$ and 33.3$\times$ faster training convergence compared to DiT and LightningDiT, respectively.
    \item We validate the effectiveness and generalizability of SFD by integrating it into existing synchronous diffusion models like ReDi and VA-VAE, thus improving their performance.
\end{itemize}

%% file: sec/2_relatedwork.tex
\section{Related Work}
\subsection{Diffusion Models for Image Generation }

Probabilistic diffusion models synthesize images by iteratively denoising from Gaussian noise. Early models~\cite{ddpm, ddim} operate in pixel space, suffering from high computational cost and slow convergence. Latent Diffusion Models (LDMs)~\cite{ldm} mitigate this by performing diffusion in a VAE-compressed latent space, greatly improving efficiency and visual fidelity.
Building upon this foundation, DiT~\cite{dit} and SiT~\cite{sit} replaced the U-Net backbone~\cite{unet} with Vision Transformers, demonstrating superior scalability and generative capacity. More recent efforts have sought to accelerate convergence by optimizing the diffusability of latent representations. For instance, \cite{Diffusability} regularizes the frequency spectrum of the latent space to make it more compatible with diffusion dynamics. Despite these advances, standard LDMs still treat all latent components uniformly during denoising, leaving the coarse-to-fine nature~\cite{dctdiff} of the diffusion synthesis process implicit.
In contrast, SFD explicitly models this hierarchical evolution via asynchronous denoising: high-level semantic components are denoised earlier and progressively guide the refinement of low-level textures at cleaner noise levels, thereby accelerating convergence and improving generation quality.

\subsection{Semantic Representation Enhanced Diffusion} 

Motivated by the discriminative gap between generative models and pretrained visual encoders, a parallel line of research seeks to enhance diffusion models with external semantic representations. REPA~\cite{REPA} performs feature-space alignment between diffusion features and pretrained visual encoders. REPA-E~\cite{repa-e} extends this alignment by enabling end-to-end joint optimization of the VAE and diffusion model. 
REG~\cite{reg} and ReDi~\cite{redi} jointly learn the distribution of low-level VAE features and high-level semantic features from DINOv2~\cite{dinov2}, where REG employs the class token as the semantic descriptor while ReDi adopts PCA-compressed patch embeddings.
Another line of work focuses on improving or changing the VAE latent space.
VA-VAE~\cite{vavae} aligns latent space with pretrained vision foundation models to to enrich its semantic representations.
RAE~\cite{RAE} and SVG~\cite{svg} replace the conventional VAE with pretrained visual encoder representations, where RAE adapts the diffusion transformer with a wide DDT head, while SVG employs a residual branch to capture fine-grained details.
Collectively, these approaches reveal that integrating semantic signals into the diffusion process benefits generation. We extend this paradigm further with SFD, which introduces asynchronous denoising schedules that allow semantics to be established earlier and guide the refinement of low-level textures throughout generation.

\subsection{Asynchronous Denoising Methods }
Asynchronous denoising allows different components (e.g., tokens, spatial regions, pixels) to evolve under distinct noise schedules rather than enforcing strict synchronicity. Diffusion Forcing~\cite{diffusion_forcing} assigns each token an independent noise level and enables arbitrary per-token denoising schedule, combining the strengths of next-token prediction models and full-sequence diffusion models. AsynDM~\cite{asynDM} dynamically modulates per-pixel timestep schedules, allowing prompt-related regions to denoise more gradually and thereby improving text-to-image alignment.  In this paper, SFD applies asynchronous denoising to semantic and texture subspaces within the latent representation, enabling early semantic guidance during denoising while preserving a simple and unified latent diffusion architecture.

%% file: sec/3_method.tex
\section{Method}

We propose Semantic-First Diffusion (SFD), which employs asynchronous denoising to harmonize semantic and texture modeling, achieving faster convergence and superior performance without sacrificing reconstruction fidelity.
Section~\ref{subsec:preliminaries} introduces the preliminaries. Section~\ref{subsec:Composite_Latent_Construction} describes our Semantic VAE and composite latent construction, and Section~\ref{subsec:SFD} presents the complete SFD framework.

\subsection{Preliminaries }
\label{subsec:preliminaries}

Flow matching-based~\cite{flow1, flow2, flow3, sit} diffusion models learn to reverse a continuous noising process that transforms clean data into Gaussian noise. Following the flow matching formulation, the forward process is modeled as a linear interpolation:
\begin{equation}
    \mathbf{x}_t = t\mathbf{x}_1 + (1 - t) \mathbf{x}_0,
\end{equation}
where $\mathbf{x}_1 \sim p(\mathbf{x})$ denotes clean data sampled from the data distribution, $\mathbf{x}_0 \sim \mathcal{N}(\mathbf{0}, \mathbf{I})$ denotes sampled Gaussian noise, and $t \in [0,1]$ denotes time. Generation starts at $t = 0$ with random noise and follows the learned velocity field toward clean data at $t = 1$.

The evolution of $\mathbf{x}_t$ is governed by the velocity field:
\begin{equation}
    \mathbf{v}(\mathbf{x}_t, t) = \mathbb{E}[\dot{\mathbf{x}}_t \mid \mathbf{x}_t] = \mathbb{E}[\mathbf{x}_1 - \mathbf{x}_0 \mid \mathbf{x}_t].
\end{equation}

This velocity field uniquely defines the associated ordinary differential equation (ODE) or stochastic differential equation (SDE) for sampling. It is also closely related to the score function $s(\mathbf{x}_t, t)$ through a simple proportional relationship, so estimating either one is sufficient.

During training, a neural network $\mathbf{v}_{\theta}(\mathbf{x}_t, t)$ is optimized to approximate the velocity field by minimizing:
\begin{equation}
    \mathcal{L}_{\text{vel}}(\theta) = \int_{0}^{1} \mathbb{E}_{\mathbf{x}_1, \mathbf{x}_0} \left[ \|\mathbf{v}_{\theta}(\mathbf{x}_t, t) - (\mathbf{x}_1 - \mathbf{x}_0)\|_2^2 \right] \, dt.
\end{equation}
During inference, a numerical solver integrates this learned velocity field from noise to data.

\subsection{Composite Latent Construction}
\label{subsec:Composite_Latent_Construction}
\subsubsection{Semantic VAE}
\label{subsubsec:Semantic VAE}

\begin{figure}[t]
\vspace{-4mm}
  \centering
  \includegraphics[width=\linewidth]{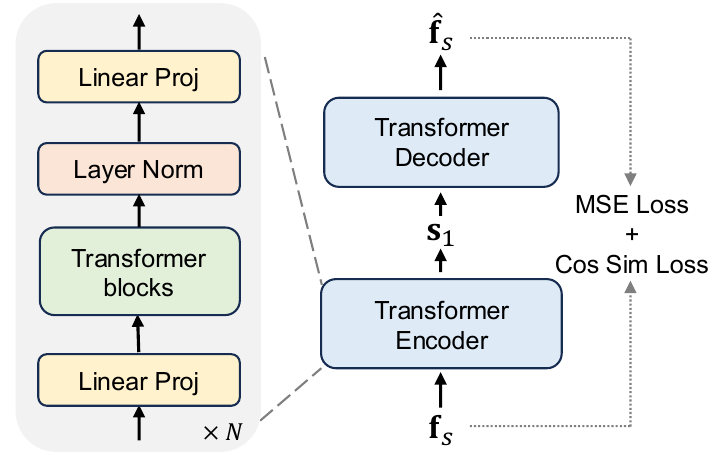}
  \vspace{-4mm}
  \caption{\textbf{Architecture of the Semantic VAE (SemVAE).} A Transformer-based VAE compresses high-dimensional vision foundation model (VFM) features into compact semantic latents.}
  \label{fig: semVAE}
  \vspace{-4mm}
\end{figure}

To comprehensively leverage high-level semantics from pretrained vision foundation models, we introduce a dedicated Semantic VAE (SemVAE) specifically designed to compress rich semantic features into compact latent representations while maintaining their spatial layout and minimizing information loss.

Figure~\ref{fig: semVAE} illustrates the architecture of SemVAE. Given an input image $\mathbf{x}_1$, we extract its patch-level semantic features 
\(
\mathbf{f}_{\text{s}} = f(\mathbf{x}_1) \in \mathbb{R}^{L \times C_\text{in}}
\) via a frozen vision foundation model (VFM) denoted as $f(\cdot)$,
where \(L\) denotes the number of flattened patches and \(C_\text{in}\) is the VFM feature dimension.
The SemVAE encoder $\mathcal{E}_s(\cdot)$, consisting of a linear projection layer, four Transformer blocks, a LayerNorm layer, and an output linear layer, maps these features to a lower-dimensional latent space:
\begin{equation}
\mathbf{h}_s = \mathcal{E}_s(\mathbf{f}_s),
\end{equation}
where \(\mathbf{h}_s \in \mathbb{R}^{L \times 2C_s}\) and \(C_s\) denotes the latent dimension. The factor of 2 accounts for the mean and variance parameters.

The encoder outputs the parameters of a Gaussian distribution. Specifically, \(\mathbf{h}_s\) is split into mean and variance:
\begin{equation}
\boldsymbol{\mu}, \boldsymbol{\sigma}^2 = \mathbf{h}_s[:, :C_s], \; \mathbf{h}_s[:, C_s:],
\end{equation}
and the latent variable $\mathbf{s}_1 \in \mathbb{R}^{L \times C_s}$ is sampled via the reparameterization trick~\cite{reparameterisation}:
\begin{equation}
\mathbf{s}_1 = \boldsymbol{\mu} + \boldsymbol{\sigma} \odot \boldsymbol{\epsilon}, \quad \boldsymbol{\epsilon} \sim \mathcal{N}(\mathbf{0}, \mathbf{I}).
\end{equation}

The SemVAE decoder $\mathcal{D}_s(\cdot)$ mirrors the encoder architecture and reconstructs the original VFM features from the latent variables:
\begin{equation}
\hat{\mathbf{f}}_s = \mathcal{D}_s(\mathbf{s}_1), \quad \hat{\mathbf{f}}_s \in \mathbb{R}^{L \times C_\text{in}}.
\end{equation}

\paragraph{Training Objective.}
The SemVAE is trained with a combination of reconstruction and regularization losses. The reconstruction quality is ensured by MSE loss and cosine similarity loss:
\begin{equation}
\mathcal{L}_{\mathrm{MSE}} = \|\hat{\mathbf{f}}_s - \mathbf{f}_s\|^2, \quad
\mathcal{L}_{\mathrm{cos}} = 1 - \frac{\hat{\mathbf{f}}_s \cdot \mathbf{f}_s}{\|\hat{\mathbf{f}}_s\| \|\mathbf{f}_s\|},
\end{equation}
where $\mathcal{L}_{\mathrm{MSE}}$ enforces reconstruction fidelity, while $\mathcal{L}_{\mathrm{cos}}$ ensures directional alignment of the feature vectors. The KL divergence regularizes the latent space:
\begin{equation}
\begin{aligned}
\mathcal{L}_{\mathrm{KL}} &= D_{\text{KL}}(q(\mathbf{s}_1|\mathbf{f}_s) \| \mathcal{N}(\mathbf{0}, \mathbf{I})) \\
&= \frac{1}{2}\sum_{i}\left(\mu_i^2 + \sigma_i^2 - \log \sigma_i^2 - 1\right).
\end{aligned}
\end{equation}
The total training loss is:
\begin{equation}
\mathcal{L}_{\mathrm{SemVAE}} = 
\mathcal{L}_{\mathrm{MSE}} + 
\mathcal{L}_{\mathrm{cos}} + 
\lambda_\text{kl}\mathcal{L}_{\mathrm{KL}}.
\end{equation}
$\lambda_\text{kl}$ is set as $10^{-7}$ by default.
Once trained, SemVAE is frozen during diffusion model training.

\subsubsection{Latent Construction}

\begin{figure}[t]
\vspace{-4mm}
  \centering
  \includegraphics[width=\linewidth]{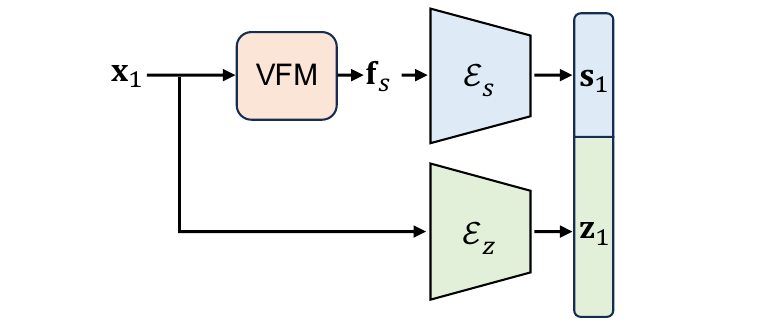}
  \vspace{-6mm}
    \caption{\textbf{Composite Latent Construction.} An input image is encoded into semantic and texture latents via distinct VAE encoders, which are then concatenated to form a composite latent for asynchronous diffusion modeling.}
  \label{fig: hyb_latent}
  \vspace{-4mm}
\end{figure}

As illustrated in Figure~\ref{fig: hyb_latent}, the composite latent is constructed by combining compressed high-level semantics $\mathbf{s}_1$ and low-level textures $\mathbf{z}_1$, which are encoded via SemVAE encoder $\mathcal{E}_s$ and texture VAE encoder $\mathcal{E}_z$, respectively. Here we implement SD-VAE~\cite{ldm} as the texture VAE. The two latents are concatenated along the channel dimension:
\begin{equation}
\mathbf{c} = [\mathbf{s}_1, \mathbf{z}_1] \in \mathbb{R}^{L \times (C_s + C_z)},
\end{equation}
where $[\cdot, \cdot]$ denotes channel-wise concatenation, and $C_s$, $C_z$ are the semantic and texture channel dimensions.

\subsection{Semantic-First Diffusion}
\label{subsec:SFD}
The core motivation of Semantic-First Diffusion (SFD) is to enable semantic latents to be denoised ahead of texture features, thereby providing clearer structural guidance throughout the asynchronous generation process. We describe its key components below.

\paragraph{Distinct timesteps for semantics and textures.}
To model semantics and textures asynchronously with a fixed temporal offset $\Delta t$ while ensuring both timesteps remain within $[0,1]$, distinct timesteps $t_s$ and $t_z$ are assigned to the semantic and texture latents during training. Specifically, for each image, we first sample the semantic timestep $t_s$ from an extended interval, then derive the texture timestep $t_z$ by subtracting the offset $\Delta t$, and finally clamp both to $[0,1]$:
\begin{align}
t_s &\sim \mathcal{U}(0,\, 1+\Delta t),\\
t_z &= \max(0,\, t_s - \Delta t),\\
t_s &= \min(t_s,\, 1),
\end{align}
which ensures $t_s, t_z \in [0,1]$ and $t_s \ge t_z$. This guarantees the semantic latent experiences less noise corruption than the texture latent at each denoising step, thereby providing clearer structural guidance for texture denoising.

\begin{figure}[t]
  \centering
  \includegraphics[width=\linewidth]{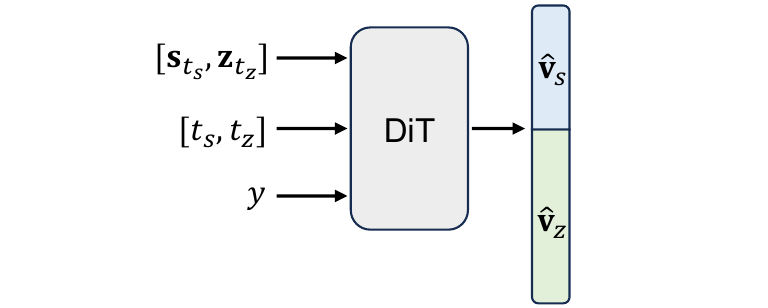}
  \vspace{-6mm}
  \caption{\textbf{Input and output of Diffusion Transformer.} A DiT backbone takes as input a composite latent that combines noisy semantic and texture features $[\mathbf{s}_{t_s}, \mathbf{z}_{t_z}]$, along with their respective timestep $[t_s, t_z]$ and class label $y$. It jointly predicts the velocities of both semantics and textures.}
  \label{fig: diffusion_train}
  \vspace{-4mm}
\end{figure}

\paragraph{Diffusion transformer with dual timesteps.}

As shown in Figure~\ref{fig: diffusion_train}, the diffusion model adopts a Transformer backbone $\mathbf{v}_\theta(\cdot)$ that takes as input the noisy composite latent $[\mathbf{s}_{t_s}, \mathbf{z}_{t_z}]$ at different noise levels, two separate timesteps $[t_s,\, t_z]$, and the class label $y$:
\begin{equation}
[\hat{\mathbf{v}}_{s}, \hat{\mathbf{v}}_{z}] =\mathbf{v}_\theta\big([\mathbf{s}_{t_s}, \mathbf{z}_{t_z}],\, [t_s,\, t_z],\, y\big),
\end{equation}
where $\hat{\mathbf{v}}_s$ and $\hat{\mathbf{v}}_z$ denote the predicted velocities of the semantic and texture components, respectively.

\vspace{-2mm}
\paragraph{Training objective.}
The training objective combines velocity prediction losses for both semantic and texture latents:
\begin{equation}
\begin{aligned}
\mathcal{L}_{\mathrm{pred}}
&= \mathbb{E}_{\mathbf{s}_0, \mathbf{s}_1, \mathbf{z}_0, \mathbf{z}_1, t_s, t_z} \Big[
\big\|
\hat{\mathbf{v}}_{z}
- (\mathbf{z}_1 - \mathbf{z}_0)
\big\|^2  \\
&\quad+ \beta \,
\big\|
\hat{\mathbf{v}}_{s}
- (\mathbf{s}_1 - \mathbf{s}_0)
\big\|^2
\Big],
\end{aligned}
\label{eq:loss_vel}
\end{equation}
where $\mathbf{s}_0 \sim \mathcal{N}(0, I)$, $\mathbf{z}_0 \sim \mathcal{N}(0, I)$ are sampled from the prior, and $\beta$ is a weighting hyperparameter.

Additionally, the representation alignment loss from REPA~\cite{REPA} is employed, which aligns the diffusion hidden states with pretrained vision encoder representations. Formally, it is defined as:
\begin{equation}
\mathcal{L}_{\mathrm{REPA}}(\psi, \phi)
:= -\mathbb{E}_{\mathbf{s}_{t_s}, \mathbf{z}_{t_z}, t_s, t_z}
\left[
\mathcal{L}_{\text{sim}}\big(\mathbf{y}_*,\, h_{\phi}(\mathbf{h}_{t})\big)
\right],
\label{eq:repa}
\end{equation}
where $\mathbf{y}_* = f(\mathbf{x}_1)$ denotes the pretrained visual encoder output, $\mathbf{h}_{t} = f_{\psi}([\mathbf{s}_{t_s},\mathbf{z}_{t_z}], [t_s, t_z])$ is the diffusion transformer encoder output, $h_{\phi}(\mathbf{h}_{t})$ projects $\mathbf{h}_{t}$ through a trainable projection head, and $\mathcal{L}_{\text{sim}}(\cdot,\cdot)$ is the alignment function. Notably, $\mathbf{y}_*$ corresponds to the representation input for SemVAE (Section~\ref{subsubsec:Semantic VAE}). Under this formulation, $\mathcal{L}_{\mathrm{REPA}}$ can be regarded as striving to reconstruct the noisy semantic latents $\mathbf{s}_{t_s}$ back to their clean representations $\mathbf{y}_*$. Compared to the original REPA, which distills the VFM's analytical capabilities, this explicit reconstruction from semantic latents offers a more tractable learning objective, thereby better preserving the integrity of semantic information and enabling more effective utilization of semantic knowledge.

The final objective becomes the following:
\begin{equation}
\mathcal{L}_{\text{total}} =
\mathcal{L}_{\text{vel}}
+ \lambda\, \mathcal{L}_{\text{REPA}}.
\label{loss}
\end{equation}

\paragraph{Three-phase denoising schedule.}
During inference, SFD employs a three-phase asynchronous denoising schedule, as illustrated in Figure~\ref{fig:teaser}(a):
\begin{enumerate}
    \item \textbf{Semantic initialization}, where $t_s \in [0, \Delta t), t_z = 0$: Only semantic latents are denoised to establish global structural guidance.
    \item \textbf{Asynchronous generation}, where $t_s \in [\Delta t, 1], t_z \in [0, 1-\Delta t)$: Both semantic and texture latents are denoised jointly yet asynchronously, with semantics advancing slightly ahead to provide clearer structural guidance for texture generation.
    \item \textbf{Texture completion}, where $t_s = 1, t_z \in [1-\Delta t, 1]$: With semantic latents fully denoised, noisy texture latents continue to refine fine-grained details.
\end{enumerate}

Formally, two binary masks 
$\mathbf{M}_s\in\{0,1\}^{B\times C_s\times H\times W}$ 
and $\mathbf{M}_z\in\{0,1\}^{B\times C_z\times H\times W}$ are introduced 
to control the denoising updates of semantic and texture latents, respectively.
According to the three-phase asynchronous denoising schedule, 
the masks $(\mathbf{M}_s, \mathbf{M}_z)$ are defined as:
\begin{equation}
[\mathbf{M}_s,\mathbf{M}_z]=
\begin{cases}
[\mathbf{1},\,\mathbf{0}], & t_s\in[0,\,\Delta t),\; t_z=0,\\[3pt]
[\mathbf{1},\,\mathbf{1}], & t_s\in[\Delta t,\,1],\; t_z\in[0,\,1-\Delta t),\\[3pt]
[\mathbf{0},\,\mathbf{1}], & t_s=1,\; t_z\in[1-\Delta t,\,1],
\end{cases}
\end{equation}
where $\mathbf{1}$ and $\mathbf{0}$ denote all-one and all-zero tensors with shapes matching $\mathbf{M}_s$ and $\mathbf{M}_z$, respectively.
The masked velocity for updating is then computed as:
\begin{equation}
\hat{\mathbf{v}}
=
\big[\mathbf{M}_s\odot\hat{\mathbf{v}}_{s}, \mathbf{M}_z\odot\hat{\mathbf{v}}_{z}\big],
\end{equation}
where $\odot$ denotes element-wise multiplication.
This mechanism explicitly controls which latents denoise at each phase, ensuring semantic latents denoise earlier to guide texture refinement continuously. By enabling asynchronous yet coordinated updates between semantic and texture latents, SFD achieves more stable optimization and naturally aligns with the coarse-to-fine generation paradigm of diffusion models.

Notably, while SFD extends the denoising timestep range by $\Delta t$, we proportionally increase the interval between successive steps, keeping the total number of diffusion steps fixed. Therefore, no additional denoising steps are required for inference. Upon completion, only the fully denoised texture latent $\mathbf{z}_1$ is decoded to the final image.

%% file: sec/4_experiments.tex
\section{Experiments}

\subsection{Experimental Setup}

\paragraph{Implementation details.}
We employ SD-VAE~\cite{ldm} f16-d32 from LightningDiT~\cite{vavae} to encode texture into 32-channel latents with $16 \times$ spatial downsampling, and the SemVAE encoder (29M parameters) to encode semantic features extracted by DINOv2-B with registers~\cite{dinov2, darcet2023vision} into 16-channel latents. 
The concatenated 48-channel representation forms a unified 256-token latent for each $256 \times 256$ image. We adopt LightningDiT~\cite{vavae} as the diffusion backbone and train on ImageNet-1K~\cite{imagenet} with a batch size of 256, a learning rate of $1 \times 10^{-4}$, and the AdamW optimizer. 
We set $\beta = 2.0$ and $\Delta t = 0.3$. For REPA, we set $\lambda = 1.0$, the alignment depth to 2, and use cosine similarity as the similarity function.
For sampling, the dopri5 solver~\cite{dormand1980family} with adaptive sampling steps is employed, following the implementation in LightningDiT\footnote{\url{https://github.com/hustvl/LightningDiT}}~\cite{vavae}. The absolute and relative tolerances are set to $10^{-6}$ and $10^{-3}$, respectively. AutoGuidance~\cite{autoguidance} is used as the guidance method with a DiT-B degradation model. Implementation details of SemVAE and complete configurations of diffusion models are provided in Appendix~\ref{secA}.

\paragraph{Evaluation protocol.}
We adopt comprehensive quantitative metrics to assess generation quality: Fr{\'e}chet Inception Distance (FID)~\cite{fid} for visual realism, structural FID (sFID)~\cite{sfid} for spatial coherence, Inception Score (IS)~\cite{IS} for class-conditional diversity, Precision (Prec.) for sample fidelity, and Recall (Rec.) for distribution coverage~\cite{recall}. All metrics are computed on 50K generated samples following the standardized ADM~\cite{ADM} evaluation pipeline.

\subsection{Main Results}

\begin{figure}[t]
  \centering
  \includegraphics[width=\linewidth]{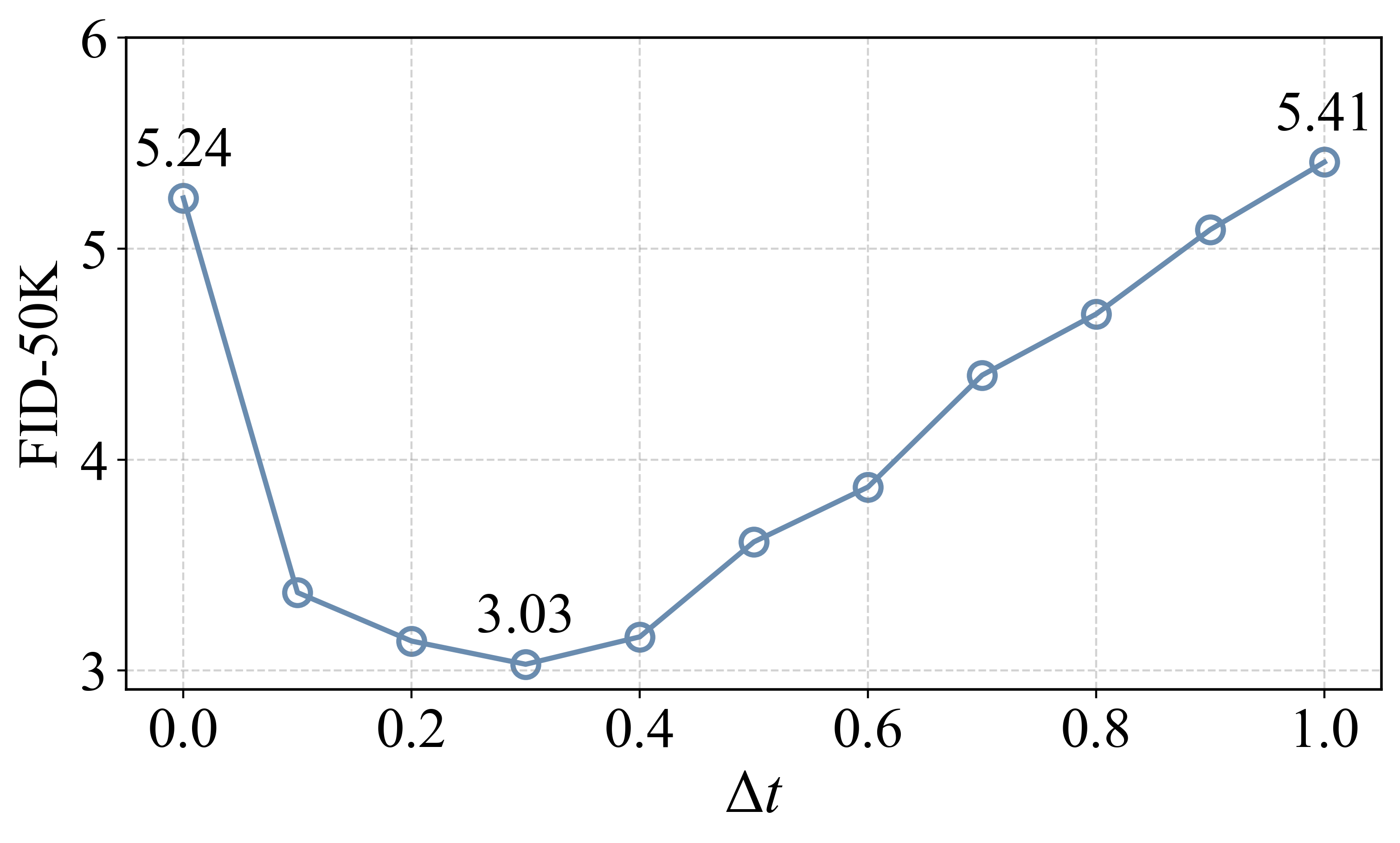}
  \vspace{-6mm}
  \caption{\textbf{Effect of the temporal offset $\Delta t$ in asynchronous denoising.} A moderate offset ($\Delta t=$ 0.3) yields the lowest FID, indicating the best semantic--texture cooperation. }
  \label{fig: fid_vs_delta_t}
  \vspace{-4mm}
\end{figure}

\paragraph{Effect of Temporal Offset in Asynchronous Denoising.}\noindent 

We analyze how the temporal offset $\Delta t$ between semantics and textures influences SFD performance. In this experiment, we set the learning rate to $2 \times 10^{-4}$ and train for 400K iterations to accelerate convergence. As shown in Figure~\ref{fig: fid_vs_delta_t}, when $\Delta t = 0$, SFD degenerates to conventional joint denoising of semantic and texture representations, similar to ReDi~\cite{redi} and REG~\cite{reg}. As $\Delta t$ increases, FID gradually decreases and reaches its optimal value of 3.03 at $\Delta t = 0.3$, corresponding to our proposed semantics-lead-texture scheme. In this setting, semantics evolve slightly ahead of textures, providing clearer global guidance while maintaining cooperative optimization.
However, increasing $\Delta t$ beyond 0.3 progressively degrades performance. When $\Delta t = 1.0$, the model reduces to a teacher-forcing sequential generation scheme, where semantics are fully synthesized before texture generation begins, leading to training-inference mismatch and suboptimal results.
Overall, these results demonstrate that a moderate offset ($\Delta t = 0.3$) achieves the optimal trade-off between early semantic stabilization and texture collaboration, better harmonizing the joint modeling of semantics and textures and thereby yielding the best generation quality.

\paragraph{Accelerating training convergence.}

Table~\ref{tab:fid_comparison} presents a comprehensive comparison between DiT~\cite{dit}, LightningDiT~\cite{vavae}, REPA~\cite{REPA}, and our SFD on ImageNet 256$\times$256 without guidance. Results for LightningDiT and its REPA variant are our reproductions. SFD consistently achieves superior FID performance while significantly accelerating convergence across all evaluated model scales.
For smaller models trained for 400K iterations, SFD reduces FID from 21.45 to 10.40 for LightningDiT-B/1 + REPA, and from 7.48 to 3.89 for LightningDiT-L/1 + REPA. In the large-scale setting, LightningDiT-XL/1 with SFD achieves FID 3.53 at only 400K iterations, outperforming LightningDiT-XL/1 with REPA at 4M iterations by 2.31 points (from 5.84 to 3.53) and vanilla DiT-XL/2 at 7M iterations by 6.09 points (from 9.62 to 3.53), with only 10\% and 5.7\% of the training cost, respectively. Notably, SFD achieves comparable performance to DiT-XL trained for 7M iterations and LightningDiT-XL/1 trained for 4M iterations in just 70K and 120K iterations, achieving 100$\times$ and 33.3$\times$ faster convergence (see Figure~\ref{fig:teaser}(b)).

\begin{table*}[t]
\centering
\begin{minipage}[t]{0.36\textwidth}
\centering
\small
\caption{\textbf{FID comparison on ImageNet 256$\times$256 without guidance} across various model sizes for DiT with REPA and SFD (ours).}
\label{tab:fid_comparison}
\begin{adjustbox}{max width=\linewidth}
\begin{tabular}{l c c c}
\toprule
Model & \#Params & Iter. & FID$\downarrow$ \\
\midrule
DiT-B/2       & 130M & 400K & 43.47 \\
LightningDiT-B/1 & 130M & 400K & 22.86 \\
+~REPA               & 130M & 400K & 21.45 \\
\rowcolor{lightblue}+~SFD (Ours) & 130M & 400K & 10.40 \\
\midrule
DiT-L/2       & 458M & 400K & 23.33 \\
LightningDiT-L/1 & 458M & 400K & 10.08 \\
+~REPA               & 458M & 400K & 7.48 \\
\rowcolor{lightblue}+~SFD (Ours) & 458M & 400K & 3.89 \\
\midrule
DiT-XL/2     & 675M & 400K & 19.47 \\
DiT-XL/2       & 675M & 7M   & 9.62 \\
LightningDiT-XL/1 & 675M & 400K & 9.29 \\
LightningDiT-XL/1 & 675M & 1M & 7.48 \\
LightningDiT-XL/1 & 675M & 2M & 6.88 \\
LightningDiT-XL/1 & 675M & 4M   & 6.50 \\
+~REPA               & 675M & 400K & 6.94 \\
+~REPA               & 675M & 1M   & 6.17 \\
+~REPA               & 675M & 2M   & 5.87 \\
+~REPA               & 675M & 4M   & 5.84 \\
\rowcolor{lightblue}+~SFD (Ours) & 675M & 70K & 8.79 \\
\rowcolor{lightblue}+~SFD (Ours) & 675M & 120K & 6.22 \\
\rowcolor{lightblue}+~SFD (Ours) & 675M & 400K & 3.53 \\
\rowcolor{lightblue}+~SFD (Ours) & 675M & 1M   & 2.82 \\
\rowcolor{lightblue}+~SFD (Ours) & 675M & 2M   & 2.74 \\
\rowcolor{lightblue}+~SFD (Ours) & 675M & 4M   & 2.54 \\
\bottomrule
\end{tabular}
\end{adjustbox}
\end{minipage}
\hfill
\begin{minipage}[t]{0.61\textwidth}
\centering
\small
\caption{\textbf{System-level comparison of class-conditional generation on ImageNet 256$\times$256 with guidance.} 
Performance metrics are annotated with $\uparrow$ (higher is better) and $\downarrow$ (lower is better).}
\label{tab:imagenet256_comparison}
\begin{adjustbox}{max width=\textwidth}
\begin{tabular}{lccccccc}
\toprule
Model & Epochs & \#Params & FID$\downarrow$ & sFID$\downarrow$ & IS$\uparrow$ & Pre.$\uparrow$ & Rec.$\uparrow$ \\
\midrule
\multicolumn{7}{c}{\textit{Autoregressive Models}} \\
\midrule
VAR~\cite{VAR} & 350 & 2.0B & 1.80 & - & \textbf{365.4} & \textbf{0.83} & 0.57 \\
MAR~\cite{MAR} & 800 & 943M & 1.55 & - & 303.7 & 0.81 & 0.62\\
xAR~\cite{xAR} & 800 & 1.1B & 1.24 & - & 301.6 & \textbf{0.83} & 0.64 \\
\midrule
\multicolumn{7}{c}{\textit{Latent Diffusion Models}} \\
\midrule
DiT-XL~\cite{dit} & 1400 & 675M & 2.27 & 4.60 & 278.2 & \textbf{0.83} & 0.57 \\
MaskDiT~\cite{maskdit} & 1600 & 675M & 2.28 & 5.67 & 276.6 & 0.80 & 0.61 \\
SiT-XL~\cite{sit} & 1400 & 675M & 2.06 & 4.50 & 270.3 & 0.82 & 0.59 \\
FasterDiT~\cite{fasterdit} & 400 & 675M & 2.03 & 4.63 & 264.0 & 0.81 & 0.60 \\
MDT~\cite{masked_diffusion(MDT)} & 1300 & 675M & 1.79 & 4.57 & 283.0 & 0.81 & 0.61 \\
MDTv2~\cite{mdtv2} & 1080 & 675M & 1.58 & 4.52 & 314.7 & 0.79 & 0.65 \\
DDT~\cite{ddt} & 400 & 675M  & 1.26 & - & 310.6 & 0.79 & 0.65 \\
\midrule
\multicolumn{7}{c}{\textit{Leveraging Visual Representations}} \\
\midrule
VA-VAE~\cite{vavae} & 800 & 675M   & 1.35 & 4.15 & 295.3 & 0.79 & 0.65 \\
REPA~\cite{REPA} & 800 & 675M & 1.42 & 4.70 & 305.7 & 0.80 & 0.65\\
REPA-E~\cite{repa-e} & 800 & 675M  & 1.12 & 4.09 & 302.9 & 0.79 & 0.66 \\
ReDi~\cite{redi} & 800 & 675M & 1.61 & 4.66 & 295.1 & 0.78 & 0.64 \\
REG~\cite{reg}  & 800 & 677M & 1.36 & 4.25 & 299.4 & 0.77 & 0.66 \\
RAE~\cite{RAE} (DiT-XL)  & 800 & 676M & 1.41 & - & 309.4 & 0.80 & 0.63 \\
RAE~\cite{RAE} ($\text{DiT}^{\text{DH}}$-XL) & 800 & 839M & 1.13 & - & 262.6 & 0.78 & \textbf{0.67} \\
\rowcolor{lightblue}SFD (XL) & 80 & 675M & 1.30 & 3.87 & 233.4 & 0.78 & 0.64 \\
\rowcolor{lightblue}SFD (XL) & 800 & 675M  & 1.06 & 3.89 & 267.0 & 0.78 & \textbf{0.67} \\
\rowcolor{lightblue}SFD (XXL)& 80 & 1.0B & 1.19 & 4.00 & 240.4 & 0.78 & 0.65 \\
\rowcolor{lightblue}SFD (XXL)& 800  & 1.0B  & \textbf{1.04} & \textbf{3.75} & 264.2 & 0.78 & 0.66 \\
\bottomrule
\end{tabular}
\end{adjustbox}
\end{minipage}
\end{table*}

\begin{figure*}
  \centering
  \includegraphics[width=\textwidth]{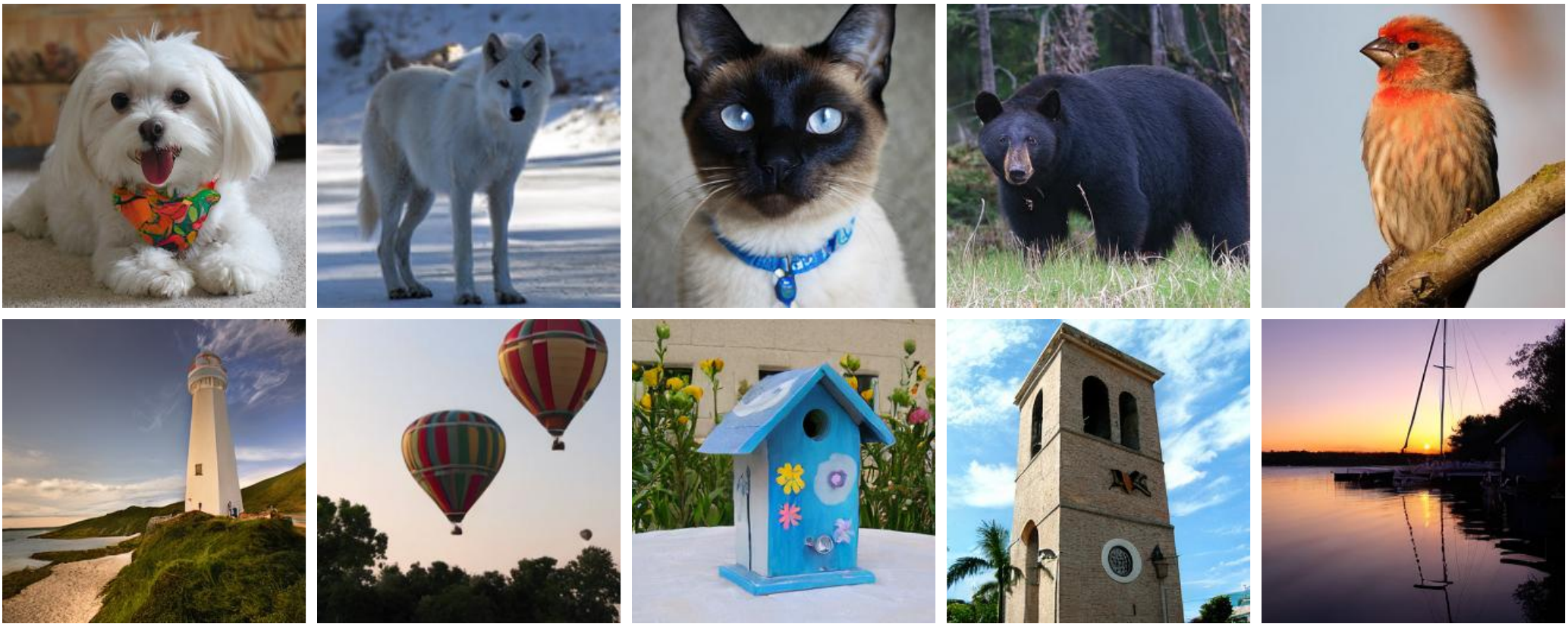}
  \vspace{-4mm}
  \caption{\textbf{Qualitative samples from our model trained at 256$\times$256 resolution.}}
  \label{fig:qualitative_samples}
  \vspace{-4mm}
\end{figure*}

\paragraph{Comparison with SOTA methods.} 

Table~\ref{tab:imagenet256_comparison} presents a system-level comparison with recent state-of-the-art methods with guidance. Our proposed SFD achieves both significantly faster convergence and superior generation performance. Remarkably, SFD surpasses DiT-XL trained for 1400 epochs within only 80 epochs, demonstrating exceptional training efficiency. At this early stage (80 epochs), SFD already achieves impressive FID scores of 1.30 with LightningDiT-XL and 1.19 with the 1.0B-parameter LightningDiT-XXL, both surpassing many existing methods. With extended training to 800 epochs, SFD achieves new state-of-the-art results of FID 1.06 with LightningDiT-XL and FID 1.04 with LightningDiT-XXL on ImageNet 256$\times$256. Complete comparisons with and without guidance are provided in Appendix~\ref{secB}.

\subsection{Ablation Studies}
All ablation experiments are conducted using the LightningDiT-XL model trained for 400K iterations on ImageNet 256$\times$256. Models are optimized with the AdamW optimizer using a learning rate of $2\times10^{-4}$ and $\beta_2 = 0.95$. FID-50K is reported as the evaluation metric.

\begin{table}[t]
\centering
\caption{\textbf{Ablation study on different components of SFD.}}
\label{tab:ablation_repa_semvae_semfirst}
\begin{tabular}{cccc}
\toprule 
REPA & SemVAE & Semantic-First & FID↓ \\
\midrule
\ding{55} & \ding{55} & \ding{55} & 8.17 \\
\ding{51} & \ding{55} & \ding{55} & 7.08 \\
\ding{51} & \ding{51} & \ding{55} & 5.24 \\
\rowcolor{lightblue}\ding{51} & \ding{51} &\ding{51} & \textbf{3.03} \\
\bottomrule
\end{tabular}
\end{table}

\vspace{-4mm}
\paragraph{Effect of different components.}
Table~\ref{tab:ablation_repa_semvae_semfirst} summarizes the contribution of each component in SFD. Starting from the baseline with FID 8.17, adding REPA yields moderate improvement to FID 7.08. Introducing semantic latents from SemVAE substantially improves performance to FID 5.24, confirming the effectiveness of explicit semantic representations. Finally, integrating the semantic-first mechanism further reduces FID to 3.03, demonstrating the effectiveness of our proposed asynchronous denoising strategy.

\paragraph{Ablation on semantic latent compression methods.}
Table~\ref{tab:semenc_semfirst} compares PCA dimensionality reduction employed in ReDi with our SemVAE as different compression methods. SemVAE achieves a significantly superior FID of 3.03 compared to PCA's 4.06, demonstrating the necessity of preserving semantic information completeness.

\paragraph{Other Ablations.} 
Due to space limitations, additional ablation studies on vision foundation model selection and scaling, SemVAE bottleneck dimension, semantic loss weight $\beta$, and REPA parameters are deferred to Appendix~\ref{secC}.

\begin{table}[t]
\centering
\begin{minipage}{0.48\linewidth}
\centering
\caption{\textbf{Ablation on semantic latent compression methods.}}
\label{tab:semenc_semfirst}
\begin{tabular}{lc}
\toprule
Method & FID$\downarrow$ \\
\midrule
PCA & 4.06 \\
\rowcolor{lightblue}SemVAE & \textbf{3.03} \\
\bottomrule
\end{tabular}
\end{minipage}
\hfill
\begin{minipage}{0.48\linewidth}
\centering
\caption{\textbf{Semantic-First helps with ReDi~\cite{redi}.}}
\label{tab:semfirst_for_redi}
\begin{tabular}{ccc}
\toprule
Semantic-First  & FID$\downarrow$ \\
\midrule
\ding{55} & 5.33 \\
\rowcolor{lightblue}\ding{51} & \textbf{4.41} \\
\bottomrule
\end{tabular}
\end{minipage}
  \vspace{-4mm}
\end{table}

\subsection{Generalization of Semantic-First Mechanism}
\label{sec: sfd_For_redi}
To validate the generalization of our semantic-first mechanism to other methods, we conduct experiments on ReDi~\cite{redi} and VA-VAE~\cite{vavae}. ReDi employs PCA-reduced DINOv2-B features as semantic latents and concatenates them with texture latents encoded by SD-VAE for simultaneous denoising. As shown in Table~\ref{tab:semfirst_for_redi}, incorporating our semantic-first mechanism into ReDi improves FID from 5.33 to 4.41, demonstrating the effectiveness of the semantic-first approach for methods that leverage semantic-texture composition. Due to space constraints, results on VA-VAE are provided in Appendix~\ref{secB}.

\subsection{Reconstruction Performance Analysis}

\begin{figure}[t]
  \centering
  \includegraphics[width=\linewidth]{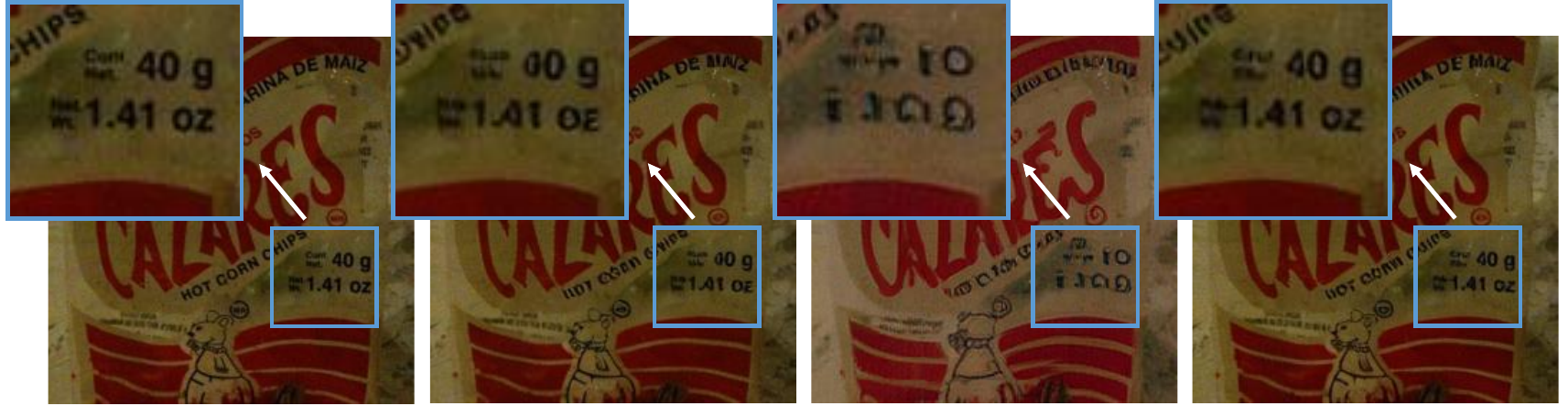}
  \vspace{-2mm}
  \caption{\textbf{Qualitative reconstruction comparison among different latent space variants.} From left to right: Original image, VA-VAE, RAE and SD-VAE reconstructions. The top-left insets show zoomed-in regions focusing on text details (``40 g''). SD-VAE achieves the best reconstruction fidelity. }
  \vspace{-4mm}
  \label{fig: compare_vaes}
\end{figure}

While recent work has highlighted the dilemma between reconstruction and generation in VAE~\cite{vavae}, our SFD framework achieves state-of-the-art generation performance, without sacrificing the reconstruction fidelity. Here we compare the reconstruction quality of latent space variants (VA-VAE~\cite{vavae} and RAE~\cite{RAE}) to SD-VAE, which is adopted as the texture VAE in our method. 
As shown in Table~\ref{tab:recon_metrics}, SD-VAE achieves the best performance with the lowest rFID of 0.26 and LPIPS of 0.089, along with the highest PSNR of 28.59 and SSIM of 0.80. VA-VAE, which aligns its latent space with visual foundation model features, enhances semantic representation but slightly compromises pixel-level reconstruction fidelity. In contrast, RAE builds its latent space purely on pretrained visual encoders, leading to texture-deficient representations and significantly degraded reconstruction quality: rFID 0.57, PSNR 18.86, LPIPS 0.256, and SSIM 0.42. 
As illustrated in Figure~\ref{fig: compare_vaes}, RAE severely loses fine-grained details such as the ``40g'' text region and exhibits noticeable color distortion, while VA-VAE shows slightly reduced fidelity and SD-VAE faithfully maintains superior reconstruction quality. By adopting SD-VAE for texture modeling while introducing a separate semantic pathway, our composite latent design enables SFD to achieve significant convergence acceleration and superior generation performance without sacrificing reconstruction fidelity. This preservation of reconstruction quality makes SFD inherently more suitable for complex image synthesis tasks, such as text-to-image generation and consistency-demanding image editing.

\begin{table}[t]
\centering
\caption{\textbf{Comparison of reconstruction performance.}}
\label{tab:recon_metrics}
\begin{tabular}{lcccc}
\toprule
Method & rFID$\downarrow$ & PSNR$\uparrow$ & LPIPS$\downarrow$ & SSIM$\uparrow$ \\
\midrule
VA-VAE        & 0.28 & 27.96 & 0.096 & 0.79 \\
RAE           & 0.57 & 18.86 & 0.256 & 0.42 \\
\rowcolor{lightblue}SD-VAE & \textbf{0.26} & \textbf{28.59} & \textbf{0.089} & \textbf{0.80} \\
\bottomrule
\end{tabular}

\end{table}

%% file: sec/5_conclusion.tex
\section{Conclusion}

We propose Semantic-First Diffusion (SFD), a novel paradigm that performs asynchronous denoising on semantic and texture latents in latent diffusion models. By prioritizing semantic denoising to guide texture refinement, SFD achieves faster convergence and superior generation quality. Extensive experiments on ImageNet class-conditional generation demonstrate that SFD consistently outperforms competing methods. Our findings suggest that controlling the relative denoising pace between semantics and textures is crucial for efficient generative modeling, establishing representation-level asynchronous denoising as a promising direction for future diffusion research.

%% file: sec/X_suppl.tex
\clearpage
\setcounter{page}{1}
\maketitlesupplementary

\appendix
\renewcommand{\thesection}{\Alph{section}}

\section{Additional Implementation Details}
\label{secA}

\subsection{SemVAE Configuration}

For semantic representation extraction, we employ DINOv2-B with registers~\cite{dinov2, darcet2023vision} on $256 \times 256$ images. The SemVAE architecture consists of 4 transformer blocks for both the encoder and decoder, with 29M parameters each (58M total). We train on ImageNet-1K~\cite{imagenet} for 1M iterations with random cropping as data augmentation. Detailed hyperparameters are shown in Table~\ref{tab:semvae_config}.

\subsection{Diffusion Model Configuration}

We adopt LightningDiT~\cite{vavae} as our diffusion backbone, which is available in multiple scales (B/L/XL/XXL). For latent construction, SD-VAE~\cite{ldm,vavae} (implemented in LightningDiT\footnote{\url{https://github.com/hustvl/LightningDiT}}~\cite{vavae}) encodes textures into 32 channels with 16$\times$ spatial compression, while SemVAE extracts 16-channel semantic representations. Their concatenation forms a unified 48-channel, 256-token latent for each 256$\times$256 image.

Following the ADM~\cite{ADM} preprocessing pipeline, all images are center cropped and resized to 256$\times$256 resolution. Training is conducted on ImageNet-1K~\cite{imagenet} for 800 epochs with a batch size of 256, using AdamW optimizer with a learning rate of $1\times10^{-4}$ and $\beta$ values of (0.9, 0.999). We employ logit-normal timestep sampling following LightningDiT~\cite{vavae}. For sampling, the dopri5 solver~\cite{dormand1980family} with adaptive step size is employed, with absolute and relative tolerances set to $10^{-6}$ and $10^{-3}$ respectively. Detailed hyperparameters across different model scales are shown in Table~\ref{tab:hyperparam_settings}.

\begin{table}[t]
\centering
\small
\caption{\textbf{SemVAE training configuration.}}
\label{tab:semvae_config}
\begin{tabular}{lc}
\toprule
\textbf{Component} & \textbf{Setting} \\
\midrule
\textbf{Feature Extraction} & \\
Feature Extractor & DINOv2-B-reg \\
Input Patch Size & 256$\times$256 \\
\midrule
\textbf{Architecture} & \\
Total Parameters & 58M \\
Encoder Parameters & 29M \\
Decoder Parameters & 29M \\
Encoder Blocks & 4 \\
Decoder Blocks & 4 \\
Hidden Dimension & 768 \\
Attention Heads & 6 \\
Bottleneck Channels & 16 \\
KL Weight $\lambda_{\text{kl}}$ & $10^{-7}$ \\
\midrule
\textbf{Training} & \\
Dataset & ImageNet-1K \\
Total Iterations & 1,000,000 \\
Batch Size & 64 \\
Data Augmentation & Random cropping \\
\midrule
\textbf{Optimization} & \\
Optimizer & AdamW \\
Learning Rate & $5\times10^{-5}$ \\
$(\beta_1, \beta_2)$ & (0.9, 0.999) \\
\midrule
\textbf{LR Schedule} & \\
Warmup Steps & 500 \\
Constant Steps & 800,000 \\
Annealing & Cosine to $5\times10^{-6}$ \\
\bottomrule
\end{tabular}
\end{table}

\begin{table*}[t]
\centering
\small
    \vspace{-3mm}
\caption{\textbf{Hyperparameter settings across different model scales.}}
\label{tab:hyperparam_settings}
\begin{tabular}{lcccc}
\toprule
\textbf{Backbone} & LightningDiT-B & LightningDiT-L & LightningDiT-XL & LightningDiT-XXL \\
\midrule
\textbf{Architecture} & & & & \\
\#Params & 130M & 458M & 675M & 1.0B \\
Input & $16\times16\times48$& $16\times16\times48$ & $16\times16\times48$ & $16\times16\times48$ \\
Layers & 12 & 24 & 28 & 32 \\
Hidden dim. & 768 & 1024 & 1152 & 1280 \\
Num. heads & 12 & 16 & 16 & 16 \\
\midrule
\textbf{SFD settings} & & & & \\
$\beta$ & 2.0 & 2.0 & 2.0 & 2.0 \\
$\Delta t$ & 0.3 & 0.3 & 0.3 & 0.3 \\
REPA visual encoder & DINOv2-B-reg & DINOv2-B-reg & DINOv2-B-reg & DINOv2-B-reg \\
REPA weight $\lambda$ & 1.0 & 1.0 & 1.0 & 1.0 \\
REPA alignment depth & 2 & 2 & 2 & 2 \\
REPA similarity function & cosine & cosine & cosine & cosine \\
\midrule
\textbf{Optimization} & & & & \\
Batch size & 256 & 256 & 256 & 256 \\
Optimizer & AdamW & AdamW & AdamW & AdamW \\
lr & $1\times10^{-4}$ & $1\times10^{-4}$ & $1\times10^{-4}$ & $1\times10^{-4}$ \\
$(\beta_1, \beta_2)$ & (0.9, 0.999) & (0.9, 0.999) & (0.9, 0.999) & (0.9, 0.999) \\
\midrule
\textbf{Sampling} & & & & \\
Sampler & dopri5 & dopri5 & dopri5 & dopri5 \\
Absolute tolerance & $10^{-6}$ & $10^{-6}$ & $10^{-6}$ & $10^{-6}$ \\
Relative tolerance & $10^{-3}$ & $10^{-3}$ & $10^{-3}$ & $10^{-3}$ \\
\bottomrule
\end{tabular}
\end{table*}

\begin{table*}[h]
    \centering
    \small
    \caption{\textbf{Configurations of degraded models used for guidance.}}
    \vspace{-2mm}
    \label{tab:degraded_configs}
    \begin{tabular}{lccccc}
    \toprule
    Model & Epochs & Params & Degraded Model & Iterations & Guidance Scale \\
    \midrule
    LightningDiT-XL  & 80  & 675M & LightningDiT-B & 70K  & 1.6 \\
    LightningDiT-XL  & 800 & 675M & LightningDiT-B & 70K  & 1.5 \\
    LightningDiT-XXL & 80  & 1.0B & LightningDiT-B & 60K  & 1.5 \\
    LightningDiT-XXL & 800 & 1.0B & LightningDiT-B & 120K & 1.5 \\
    \bottomrule
    \end{tabular}
\end{table*}

\subsection{Dual Timestep Embedding}
To support asynchronous denoising, SFD employs two independent timestep 
embedders corresponding to the semantic and texture timesteps $t_s$ and $t_z$.
Unlike LightningDiT~\cite{vavae}, which uses a single MLP-based embedder of hidden dimension
$H$, SFD constructs two smaller embedders whose hidden dimensions are reduced to
$H/2$. Each embedder independently processes its respective timestep, the two embeddings are then concatenated along the channel dimension and injected into 
the backbone. This design allows the model to supply distinct timestep signals to the semantic and texture latents:
\begin{equation}
\label{dual_timestep}
\mathbf{e} = [\, \tau_s(t_s),\; \tau_z(t_z) \,],
\end{equation}
where $[\,\cdot,\cdot\,]$ denotes channel-wise concatenation, and 
$\tau_s(\cdot)$ and $\tau_z(\cdot)$ are the semantic and texture timestep 
embedders.

\subsection{Evaluation Details}

\paragraph{AutoGuidance. }

We employ AutoGuidance~\cite{autoguidance} as our primary guidance method. 
Unlike Classifier-Free Guidance (CFG)~\cite{cfg}, which relies on an unconditional model, 
AutoGuidance guides the main diffusion model using a \emph{weaker version} of itself—typically a model with smaller capacity or an earlier training snapshot. 
This self-guidance mechanism effectively suppresses out-of-manifold samples by aligning the denoising trajectory toward regions of higher data density, 
thereby improving image quality without sacrificing sample diversity. 
In practice, we use the degraded LightningDiT-B model as the guiding network.
After searching, configurations of degraded models are illustrated in Tab.~\ref{tab:degraded_configs}.

\paragraph{Class-balanced Sampling. }
RAE~\cite{RAE} shows that class-balanced sampling yields more reliable and lower FID estimates. To ensure fair comparison with prior work~\cite{VAR,MAR,xAR,ddt,RAE}, we follow this protocol and adopt class-balanced sampling for FID-50K evaluation. Specifically, we generate 50 images per class (50,000 in total).

\section{Additional Experimental Results}
\label{secB}
\subsection{Complete Comparisons}

Table~\ref{tab:imagenet256_comparison_supp} presents a system-level comparison of class-conditional generation on ImageNet $256 \times 256$.
In the guidance setting, our SFD achieves state-of-the-art performance, surpassing existing methods in both FID and sFID. Notably, our SFD-XL (675M) outperforms the previous best model, RAE DiT$^\text{DH}$~(839M), with a lower FID (1.06 vs. 1.13), demonstrating superior generation quality with fewer parameters. Scaling up to SFD-XXL (1.0B) further pushes the performance boundary to a FID of 1.04.
Notably, SFD achieves a superior sFID of 3.75, outperforming previous methods by a substantial margin. Since sFID serves as a metric for structural coherence and spatial alignment, this improvement validates the advantage of our explicit compression of semantic representations with spatial layouts, which ensures robust global structure before texture refinement.

Regarding the unguided setting, SFD remains competitive but exhibits limitations in texture convergence. This is primarily attributed to the high complexity of the texture latents. Unlike methods such as ReDi~\cite{redi} or REG~\cite{reg} that utilize a standard f8d4 VAE, we employ the f16d32 variant (following LightningDiT), which results in a latent space with double the dimensionality. Consequently, modeling these high-dimensional texture latents is inherently more challenging and harder to converge.

\newcolumntype{Y}{>{\centering\arraybackslash}X}
\begin{table*}[t]
\centering
\small
\caption{\textbf{System-level comparison of class-conditional generation on ImageNet 256$\times$256.}}
\label{tab:imagenet256_comparison_supp}
\begin{tabularx}{\textwidth}{lccYYYYYYYYYYY}
\toprule
\multirow{2}{*}{\textbf{Method}} & 
\multirow{2}{*}{\textbf{Epochs}} & 
\multirow{2}{*}{\textbf{\#Params}} & 
\multicolumn{5}{c}{\textbf{Generation@256 w/o guidance}} & 
\multicolumn{5}{c}{\textbf{Generation@256 w/ guidance}} \\
\cmidrule(lr){4-8} \cmidrule(lr){9-13}
 &  &  & FID$\downarrow$ & sFID$\downarrow$ & IS$\uparrow$ & Prec.$\uparrow$ & Rec.$\uparrow$ 
 & FID$\downarrow$ & sFID$\downarrow$ & IS$\uparrow$ & Prec.$\uparrow$ & Rec.$\uparrow$ \\
\midrule
\multicolumn{13}{l}{\textbf{\textit{Autoregressive}}} \\
\midrule
VAR~\cite{VAR} & 350 & 2.0B & - & - & - & - & - & 1.80 & - & \textbf{365.4} & \textbf{0.83} & 0.57 \\  
MAR~\cite{MAR} & 800 & 943M & 2.35 & - & 227.8 & 0.79 & 0.62 & 1.55 & - & 303.7 & 0.81 & 0.62\\
xAR~\cite{xAR} & 800 & 1.1B & - & - & - & - & - & 1.24 & - & 301.6 & \textbf{0.83} & 0.64 \\
\midrule
\multicolumn{13}{l}{\textbf{\textit{Pixel Diffusion}}} \\
\midrule
ADM~\cite{ADM} & 400 & 554M & 10.94 & 6.02 & 101.0 & 0.69 & 0.63 & 3.94 & 6.14 & 215.8 & \textbf{0.83} & 0.53 \\
RIN~\cite{RIN} & 480 & 410M & 3.42 & -  & 182.0 & - & - & - & - & - & - & - \\
PixelFlow~\cite{pixelflow} & 320 & 677M & - & - & - & - & - & 1.98 & 5.83  & 282.1 & 0.81 & 0.60 \\
PixNerd~\cite{pixnerd} & 160 & 700M & - & - & - & - & - & 2.15 & 4.55 & 297.0 & 0.79 & 0.59 \\
SiD2~\cite{sid2} & 1280 & - & - & - & - & - & - & 1.38 & - & - & - & - \\
\midrule
\multicolumn{13}{l}{\textbf{\textit{Latent Diffusion}}} \\
\midrule
DiT~\cite{dit} & 1400 & 675M & 9.62 & 6.85 & 121.5 & 0.67 & 0.67 & 2.27 & 4.60 & 278.2 & \textbf{0.83} & 0.57 \\
MaskDiT~\cite{maskdit} & 1600 & 675M & 5.69 & 10.34 & 177.9 & 0.74 & 0.60 & 2.28 & 5.67 & 276.6 & 0.80 & 0.61 \\
SiT~\cite{sit} & 1400 & 675M & 8.61 & 6.32 & 131.7 & 0.68 & 0.67 & 2.06 & 4.50 & 270.3 & 0.82 & 0.59 \\
FasterDiT~\cite{fasterdit} & 400 & 675M & 7.91 & 5.45 & 131.3 & 0.67 & 0.69 & 2.03 & 4.63 & 264.0 & 0.81 & 0.60 \\
MDT~\cite{masked_diffusion(MDT)} & 1300 & 675M & 6.23 & 5.23 & 143.0 & 0.71 & 0.65 & 1.79 & 4.57 & 283.0 & 0.81 & 0.61 \\
MDTv2~\cite{mdtv2} & 1080 & 675M & - &  & - & - & - & 1.58 & 4.52 & 314.7 & 0.79 & 0.65 \\
DDT~\cite{ddt} & 400 & 675M & 6.27 & - & 154.7 & 0.68 & \textbf{0.69} & 1.26 & - & 310.6 & 0.79 & 0.65 \\
\midrule
\multicolumn{13}{l}{\textbf{\textit{Leveraging Visual Representations}}} \\
\midrule
VA-VAE~\cite{vavae} & 800 & 675M & 2.17 & 4.36 & 205.6 & 0.77 & 0.65 & 1.35 & 4.15 & 295.3 & 0.79 & 0.65 \\
REPA~\cite{REPA} & 800 & 675M & 5.90 & - & - & - & - & 1.42 & 4.70 & 305.7 & 0.80 & 0.65\\
REPA-E~\cite{repa-e} & 800 & 675M & 1.69 & \textbf{4.17} & 219.3 & 0.77 & 0.67 & 1.12 & 4.09 & 302.9 & 0.79 & 0.66 \\
ReDi~\cite{redi} & 800 & 675M & 3.30 & 4.80 & 188.9 & 0.74 & 0.68  & 1.61 & 4.66 & 295.1 & 0.78 & 0.64 \\
REG~\cite{reg} & 800 & 677M & 1.80 & 4.59 & 230.8 & 0.77 & 0.66 & 1.36 & 4.25 & 299.4 & 0.77 & 0.66 \\
RAE~\cite{RAE} (DiT-XL) & 800 & 676M & 1.87 & - & 209.7 & \textbf{0.80} & 0.63 & 1.41 & - & 309.4 & 0.80 & 0.63 \\
RAE~\cite{RAE} ($\text{DiT}^{\text{DH}}$-XL) & 800 & 839M & \textbf{1.51} & - & \textbf{242.9} & 0.79 & 0.63 & 1.13 & -  & 262.6 & 0.78 & \textbf{0.67} \\
\rowcolor{lightblue}SFD (XL) & 80 & 675M & 3.43 & 4.34 & 162.0 & 0.75 & 0.65 & 1.30 & 3.87 & 233.4 & 0.78 & 0.64\\
\rowcolor{lightblue}SFD (XL) & 800 & 675M & 2.54 & 4.38 & 191.7 & 0.75 & 0.67 & 1.06 & 3.89 & 267.0 & 0.78 & \textbf{0.67} \\
\rowcolor{lightblue}SFD (XXL)& 80 & 1.0B & 2.84  & 4.25 & 172.6 & 0.75 & 0.65 & 1.19 & 4.00 & 240.4 & 0.78 & 0.65 \\
\rowcolor{lightblue}SFD (XXL)& 800 & 1.0B & 2.38 & 4.37 & 197.9 & 0.75 & 0.67 & \textbf{1.04} & \textbf{3.75} & 264.2 & 0.78 & 0.66 \\
\bottomrule
\end{tabularx}
\end{table*}

\begin{table*}[t]
\centering
\small
\vspace{-1mm}
\caption{\textbf{FID ($\downarrow$) comparison across inference steps.} 
All models are trained for 400K iterations and evaluated using the Euler sampler without guidance. 
Reported values are FID-10K scores computed at different inference step counts.}
\label{tab:fid_inference_steps}
\begin{tabular}{lcccccc}
\toprule
\multirow{2}{*}{\textbf{Method}} & 
\multicolumn{6}{c}{\textbf{Inference steps}} \\
\cmidrule(lr){2-7}
 & 250 & 200 & 150 & 100 & 80 & 60  \\
\midrule
LightningDiT & 12.50 & 12.58 & 12.67 & 12.91 & 13.03 & 13.40  \\
LightningDiT+REPA & 10.00 & 10.10 & 10.23 & 10.50 & 10.67 & 10.94  \\
LightningDiT+VA-VAE & 7.66 & 7.66 & 7.68 & 7.70 & 7.76 & 7.83  \\
LightningDiT+ReDi & 8.58 & 8.63 & 8.72 & 8.86 & 9.02 & 9.32  \\
\rowcolor{lightblue}LightningDiT+SFD (Ours) & 6.32 & 6.26 & 6.41 & 6.35 & 6.81 & 6.77  \\
\bottomrule
\end{tabular}
\vspace{-1mm}
\end{table*}

\begin{table*}[t]
\centering
\vspace{-1mm}
\caption{\textbf{FID ($\downarrow$) comparison across inference steps} for SFD (XL) and SFD (XXL) models at 4M training iterations with guidance. Reported values are FID-50K scores.}
\label{tab:fid_steps_euler}
\small
\begin{tabular}{lcccccccccccc}
\toprule
\textbf{Method} & \textbf{dopri5} & \textbf{250} & \textbf{200} & \textbf{150} & \textbf{100} & \textbf{80} & \textbf{60} & \textbf{50} & \textbf{40} & \textbf{30} & \textbf{25} \\
\midrule
SVG~\cite{svg}
& - & - & - & - & - & - & - & - & - & - & 1.920 \\
SFD (XL) 
& 1.064 & 1.051 & 1.050 & 1.048 & 1.045
& 1.086 & 1.102 & 1.206 & 1.510 & 1.447 & 1.865 \\

SFD (XXL)
& 1.040 & 1.035 & 1.040 & 1.041 & 1.058
& 1.080 & 1.106 & 1.190 & 1.429 & 1.456 & 1.844 \\

\bottomrule
\end{tabular}
\vspace{-1mm}
\end{table*}

\begin{table*}[t]
\centering
\small
\vspace{-1mm}
\caption{\textbf{Comparison of class-random sampling and class-balanced sampling.}}
\label{tab:sampling_comparison}
\begin{tabular}{lcccccccccc}
\toprule
\multirow{2}{*}{\textbf{Method}} & 
\multicolumn{5}{c}{\textbf{Random sampling}} & 
\multicolumn{5}{c}{\textbf{Balanced sampling}} \\
\cmidrule(lr){2-6} \cmidrule(lr){7-11}
 & FID$\downarrow$ & sFID$\downarrow$ & IS$\uparrow$ & Prec.$\uparrow$ & Rec.$\uparrow$ 
 & FID$\downarrow$ & sFID$\downarrow$ & IS$\uparrow$ & Prec.$\uparrow$ & Rec.$\uparrow$ \\
\midrule
SiT~\cite{sit} & 2.06 & 4.50 & 270.3 & \textbf{0.82} & 0.59 & 1.95 & - & 259.5 & - & - \\
REPA~\cite{REPA} & 1.42 & 4.70 & 305.7 & 0.80 & 0.65 & 1.29 & - & 306.3 & \textbf{0.79} & 0.64 \\
REPA-E~\cite{repa-e} & 1.26 & 4.11 & \textbf{314.9} & 0.79 & 0.66 & 1.12 & 4.09 & 302.9 & \textbf{0.79} & 0.66 \\
DDT~\cite{ddt} & 1.40 & - &  303.6 & - & - & 1.26 & - & \textbf{310.6} & \textbf{0.79} & 0.65 \\
VA-VAE~\cite{vavae} & 1.35 & 4.15 & 295.3 & 0.79 & 0.65 & 1.23 & 4.20 & 296.0 & \textbf{0.79} & 0.65 \\
ReDi~\cite{redi}   & 1.61 & 4.66 & 295.1 & 0.78 & 0.64 & 1.60 & 5.99 & 294.7  & 0.78 & 0.64 \\
REG~\cite{reg}    & 1.36 & 4.25 & 299.4 & 0.77 & 0.66 & 1.19 & 4.44 & 305.4  & 0.78 & 0.66 \\
RAE~\cite{RAE} ($\text{DiT}^{\text{DH}}$-XL) & 1.28 & - & 262.9 & - & - & 1.13 & - & 262.6& 0.78 & \textbf{0.67} \\
\rowcolor{lightblue}SFD (XL) & \textbf{1.18} & \textbf{3.89} & 266.8 & 0.78 & \textbf{0.67} & \textbf{1.06} & \textbf{3.89} & 267.0  & 0.78 & \textbf{0.67} \\
\bottomrule
\end{tabular}
\vspace{-1mm}
\end{table*}

\subsection{Inference Strategies}

\paragraph{Inference Steps. }
Table~\ref{tab:fid_inference_steps} illustrates FID scores without guidance of various sampling steps, showing that SFD maintains strong performance even with significantly fewer inference steps.
While other baselines require 200–250 steps to approach their optimal FID, SFD already achieves a competitive score of 6.35 at only 100 steps, and further increasing the steps to 250 yields only marginal improvement (6.32). 
This observation suggests that the semantic-first design may facilitate more efficient sampling: by stabilizing global semantics early, the model requires fewer refinement steps to reach high-quality solutions.

Table~\ref{tab:fid_steps_euler} presents the FID scores with guidance across varying sampling steps. Notably, SFD (XL) achieves a superior FID of \textbf{1.045} at only 100 steps using the Euler sampler, surpassing the result yielded by the dopri5 sampler (1.064). Furthermore, in the few-step regime (25 steps), SFD (XL) maintains its advantage over SVG~\cite{svg}, recording an FID of 1.865 compared to 1.920 by SVG.

\paragraph{Class-balanced Sampling.} To ensure a rigorous comparison, we re-evaluate prior state-of-the-art methods employing the same class-balanced protocol as discussed in RAE~\cite{RAE}. Specifically, results for SiT~\cite{sit}, REPA~\cite{REPA}, and DDT~\cite{ddt} are adopted from RAE~\cite{RAE}, while REPA-E~\cite{repa-e} figures are sourced from its original publication. Additionally, we conduct independent evaluations for VA-VAE~\cite{vavae}, ReDi~\cite{redi}, and REG~\cite{reg}. The quantitative comparison results are presented in Table~\ref{tab:sampling_comparison}. 
As observed, our proposed SFD (XL) demonstrates consistent superiority across both protocols. 
Remarkably, whether using class-balanced or class-random sampling, SFD achieves the best performance in terms of FID and sFID metrics, surpassing all competing state-of-the-art methods.

\begin{table}[t]
\centering
\small
\setlength{\tabcolsep}{4pt}
\caption{\textbf{Comparison of unconditional generation on ImageNet 256$\times$256.} RG and AG are short of Representation Guidance~\cite{redi} and AutoGuidance~\cite{karras2024guiding}.}
\label{tab:imagenet_main}
\begin{tabular}{lcccc}
\toprule
Method & Epochs & Params & FID$\downarrow$ & IS$\uparrow$ \\
\midrule
DiT-XL~\cite{dit} & 400 & 675M & 30.68      & 32.7 \\
ReDi~\cite{redi}                         & 80  & 675M & 25.10              & -- \\
ReDi~\cite{redi}~(w/ RG)                        & 80  & 675M & 22.60              & -- \\
RAE~\cite{RAE}~(w/ AG)        & 200 & 839M & 4.96              & 123.1 \\
RCG~\cite{RCG}~(DiT-XL/2)                          & 400 & 675M & 4.89              & 143.2 \\
RCG~\cite{RCG}~(MAGE-L)                          & 800 & 502M & 3.44              & 186.9 \\
RCG-G~\cite{RCG}~(MAGE-L)                          & 800 & 502M & \textbf{2.15}              & \textbf{253.4} \\
\rowcolor{lightblue}SFD (w/o AG) & 80 & 675M & 10.24 & 78.5 \\
\rowcolor{lightblue}SFD (w/ AG)       & 80  & 675M & 3.77              & 127.9 \\
\rowcolor{lightblue}SFD (w/o AG)       & 200 & 675M & 8.46 & 89.9 \\
\rowcolor{lightblue}SFD (w/ AG)       & 200 & 675M & 2.90     & 148.5 \\
\bottomrule
\end{tabular}
\end{table}

\subsection{Unconditional Generation}

We further evaluate the proposed SFD on unconditional image generation on ImageNet 256$\times$256. 
During both training and sampling, we set the class label to 1000 (the null label).
As shown in Table~\ref{tab:imagenet_main}, SFD demonstrates remarkable performance with high training efficiency. 
Even without AutoGuidance (AG)~\cite{autoguidance}, SFD significantly surpasses ReDi (FID 25.10 $\to$ 10.24) after only 80 epochs and further improves to an FID of 8.46 after 200 epochs.  
With AG enabled, SFD achieves substantial gains, reaching FIDs of 3.77 and 2.90 at 80 and 200 epochs, respectively. 
We attribute these improvements to the asynchronous denoising mechanism of SFD, which becomes especially crucial in the unconditional setting. 
These results suggest that, without class labels as conditional guidance, smoother semantic representations are more easily modeled, thus providing accurate global structural cues for superior generation performance.

\begin{table}[t]
\centering
\small
\caption{\textbf{Effect of SFD for VA-VAE.}}
\label{tab:vae_semfirst}
\setlength{\tabcolsep}{12pt} 
\begin{tabular}{lcc}
\toprule
TexEnc & SFD & FID$\downarrow$ \\
\midrule
VA-VAE & \ding{55} & 4.52 \\
VA-VAE & \ding{51} & 4.14 \\
SD-VAE (ours) & \ding{51} & \textbf{3.03} \\
\bottomrule
\end{tabular}
\end{table}

\subsection{SFD for VA-VAE}
Tab.~\ref{tab:vae_semfirst} analyzes the impact of applying Semantic-First Diffusion (SFD) to VA-VAE~\cite{vavae}. 
For both VA-VAE and ReDi settings, the SFD implementations used for comparison are our reproduced versions.
When equipped with SFD, VA-VAE improves performance from an FID of 4.52 to 4.14, indicating that SFD is also compatible with joint semantic-texture latent space.
However, its overall performance still lags behind our SD-VAE-based SFD (FID 3.03).
This is likely because VA-VAE’s latent space inherently entangles semantics and textures, leaving limited flexibility for the asynchronous denoising mechanism to operate effectively.
In contrast, disentangling semantic and texture representations (as done in SD-VAE) allows the semantic latents to stabilize early and provide clearer global guidance for texture refinement, ultimately yielding higher generative quality.

\subsection{Computational Cost}

We evaluate the computational overhead introduced by integrating SFD into LightningDiT-XL. SFD modifies the backbone in two ways. First, it augments the latent representation with a 16-channel semantic latent, which introduces a marginal increase in backbone FLOPs. Second, SFD replaces the single timestep embedder in LightningDiT with two independent embedders that operate on the semantic and texture timesteps $t_s$ and $t_z$. As shown in equation~\ref{dual_timestep}, although two embedders are used, the total parameter count is actually smaller, since MLP parameters grow quadratically with hidden dimension. Consequently, two $(H/2)$-width MLPs contain only $0.5\times$ the parameters and FLOPs of a single $H$-width MLP.

Table~\ref{tab:cost_sfd} reports the computational cost comparison. 
SFD incurs only a negligible increase in FLOPs (less than 0.01\%) while delivering a dramatic improvement in FID at 400K iterations. This indicates that SFD achieves an extremely favorable cost--performance tradeoff with virtually no additional computational burden.

\begin{table}[t]
\centering
\small
\caption{\textbf{Computational cost and performance comparison} between 
LightningDiT and LightningDiT+SFD at 400K iterations on ImageNet 256$\times$256. 
SFD adds negligible computational overhead while delivering substantially improved generation quality.}
\label{tab:cost_sfd}
\begin{adjustbox}{max width=\linewidth}
\begin{tabular}{lccc}
\toprule
Method & \#Params (M)$\downarrow$ & GFLOPs$\downarrow$ & FID$\downarrow$ \\
\midrule
LightningDiT-XL & 683.39 & \textbf{116.479} & 9.29 \\
LightningDiT-XL + SFD & \textbf{682.77}  & 116.487 & \textbf{3.53} \\
\bottomrule
\end{tabular}
\end{adjustbox}
\end{table}

\section{Additional Ablation Studies}
\label{secC}
\begin{table*}[t]
\centering
\small
\caption{\textbf{Ablation on Semantic VAE design.} (a) compares different target representation models; (b) studies model scaling within DINOv2 family; (c) analyzes semantic channel capacity.}
\label{tab:repr_scaling_channel}
\newlength{\subtblht}
\setlength{\subtblht}{8.8em} 
\begin{subtable}[t]{0.31\textwidth}
\centering
\begin{minipage}[t][\subtblht][t]{\linewidth}\vspace{0pt}
\centering
\begin{tabular}{lc}
\toprule
Target Repr. & FID$\downarrow$ \\
\midrule
\textbf{DINOv2-B} & \textbf{3.03} \\
MAE-B & 6.29 \\
CLIP-B & 4.89 \\
SigLip-B & 4.15 \\
\bottomrule
\end{tabular}
\end{minipage}
\captionsetup{skip=4pt} 
\caption{Model comparison.}
\end{subtable}
\hfill
\begin{subtable}[t]{0.31\textwidth}
\centering
\begin{minipage}[t][\subtblht][t]{\linewidth}\vspace{0pt}
\centering
\begin{tabular}{lc}
\toprule
Target Repr. & FID$\downarrow$ \\
\midrule
DINOv2-S & 4.14 \\
DINOv2-B & 3.03 \\
\textbf{DINOv2-L} & \textbf{2.97} \\
\bottomrule
\end{tabular}
\end{minipage}
\captionsetup{skip=4pt}
\caption{Scaling comparison.}
\end{subtable}
\hfill
\begin{subtable}[t]{0.31\textwidth}
\centering
\begin{minipage}[t][\subtblht][t]{\linewidth}\vspace{0pt}
\centering
\begin{tabular}{cc}
\toprule
\#Channels & FID$\downarrow$ \\
\midrule
2  & 3.90 \\
4  & 3.67 \\
8  & 3.16 \\
\textbf{16} & \textbf{3.03} \\
\bottomrule
\end{tabular}
\end{minipage}
\captionsetup{skip=4pt}
\caption{Channel capacity.}
\end{subtable}
\end{table*}

\subsection{Semantic VAE Design}

Our Semantic VAE (SemVAE) compresses pretrained vision foundation model features into compact semantic representations. To investigate its design choices, we conduct a series of ablation studies on three key aspects: 
the choice of pretrained vision encoder, model scaling within the encoder family, and the number of output channels representing semantic capacity.

\paragraph{Different target representation and model scaling.}
Tab.~\ref{tab:repr_scaling_channel} (a) compares several pretrained vision encoders used as target representations. Among all candidates, DINOv2-B achieves the lowest FID of 3.03, outperforming MAE~\cite{MAE}, CLIP~\cite{CLIP}, and SigLip~\cite{siglip}, indicating that DINOv2 provides the most effective supervision for compact semantic latent learning. 
Tab.~\ref{tab:repr_scaling_channel} (b) studies different model scales within the DINOv2 family. Larger encoders yield better semantic guidance, with DINOv2-L achieving the best FID of 2.97. 
Notably, this finding stands in contrast to recent works like REG~\cite{reg} and RAE~\cite{RAE}, which identified DINOv2-B as the optimal choice and observed performance degradation when scaling to larger VFMs due to their increased dimensionality. Our results demonstrate the superiority of our explicit semantic compression strategy, which effectively handles high-dimensional features and unlocks the potential for further scaling with more powerful VFMs.
Considering the trade-off between performance and efficiency, we adopt DINOv2-B as the default pretrained visual encoder.

\paragraph{Channel capacity.} DINOv2-B outputs 768-dimensional features, which are compressed by the Semantic VAE into a lower-dimensional semantic latent. Tab.~\ref{tab:repr_scaling_channel} (c) investigates the impact of varying the latent channel capacity. We observe a consistent performance improvement as the number of channels increases from 2 to 16. This trend indicates that a higher channel capacity is essential for preserving the rich semantic information embedded in the original high-dimensional features. The 16-channel configuration achieves the best FID of 3.03, confirming that retaining more semantic details directly contributes to superior generation quality.

\subsection{Effect of semantic loss weight}

Tab.~\ref{tab:sem_loss_weight} analyzes the impact of the semantic loss weight $\beta$ in the velocity prediction objective. 
As the weight increases from 0.25 to 2.0, the FID score consistently decreases, indicating that stronger semantic supervision enhances training stability and generation performance. 
However, when $\beta$ becomes excessively large (e.g., 4.0 or 8.0), the performance degrades, suggesting that overemphasizing semantics suppresses texture learning and leads to loss of fine details. 
Overall, $\beta=2.0$ achieves the best balance between semantic guidance and texture refinement, yielding the lowest FID of 3.03.

\begin{table}[t]
\centering
\small
\caption{\textbf{Effect of semantic loss weight.}}
\label{tab:sem_loss_weight}
\begin{tabular}{lcccccc}
\toprule
Weight $\beta$ & 0.25 & 0.5 & 1.0 & 2.0 & 4.0 & 8.0 \\
\midrule
FID$\downarrow$ & 3.46 & 3.26 & 3.08 & \textbf{3.03} & 3.28 & 3.96 \\
\bottomrule
\end{tabular}
\end{table}

{
\setlength{\aboverulesep}{0pt}
\setlength{\belowrulesep}{0pt}
\begin{table}[t]
\centering
\small
\caption{\textbf{Ablation on REPA configurations.}
Depth of conducting REPA loss, loss weight $\lambda$, and loss type are included.}
\label{tab:repa_ablation}
\begin{tabular}{cccc}
\toprule
Depth & Weight $\lambda$ & Type & FID$\downarrow$ \\
\midrule
-- & -- & -- & 4.15 \\
\cellcolor{gray!20} 2 & 0.5 & cosine+MSE & \textbf{3.03} \\
\cellcolor{gray!20} 4 & 0.5 & cosine+MSE & 3.07 \\
\cellcolor{gray!20} 6 & 0.5 & cosine+MSE & 3.24 \\
\cellcolor{gray!20} 8 & 0.5 & cosine+MSE & 3.16 \\
\cellcolor{gray!20} 10 & 0.5 & cosine+MSE & 3.19 \\
\cellcolor{gray!20} 12 & 0.5 & cosine+MSE & 3.28 \\
\midrule
2 & \cellcolor{gray!20} 0.25 & cosine+MSE & 3.30 \\
2 & \cellcolor{gray!20} 0.5 & cosine+MSE & \textbf{3.03} \\
2 & \cellcolor{gray!20} 1.0 & cosine+MSE & 3.18 \\
2 & \cellcolor{gray!20} 2.0 & cosine+MSE & 3.25 \\
2 & \cellcolor{gray!20} 4.0 & cosine+MSE & 3.20 \\
\midrule
 2 & 0.5 & \cellcolor{gray!20} cosine & 3.16 \\
 2 & 0.5 & \cellcolor{gray!20} MSE & 3.13 \\
 2 & 0.5 & \cellcolor{gray!20} cosine+MSE & \textbf{3.03} \\
\bottomrule
\end{tabular}
\end{table}
}

\subsection{Effect of REPA configurations}
\paragraph{Alignment depth.} Tab.~\ref{tab:repa_ablation} presents a systematic study of REPA configurations. In our experiments, applying the REPA loss at shallow layers (specifically at depth 2) yields the best performance with an FID of 3.03, whereas the original REPA~\cite{REPA} reports optimal performance at depth 8. We attribute this discrepancy to the distinct role of the alignment loss in our framework.
While the original REPA operates as a distillation process that forces the diffusion model to gradually \textit{analyze and understand} the input latents, our approach utilizes the REPA loss to drive the model to \textit{decode and reconstruct} high-level semantic representations from the noisy compressed latents. Since decoding from a semantic latent is inherently a more straightforward task than analyzing semantics from scratch, our model can achieve effective alignment at much shallower layers. Consequently, early-layer alignment suffices to recover the semantic guidance, avoiding the need for deeper intervention.

\paragraph{REPA loss weight $\lambda$.} For the REPA loss weight $\lambda$, the model achieves the lowest FID at $\lambda=0.5$. This indicates that a moderate alignment strength provides a good balance between semantic consistency and generative fidelity. 

\paragraph{REPA similarity function.} We also compare results of different REPA similarity functions. 
While conventional REPA employs cosine similarity for feature alignment, we additionally explore combining cosine and MSE losses inspired by our SemVAE training. The combined objective (cosine+MSE) achieves the best performance of 3.03 FID score, outperforming single-loss variants. This suggests that employing a similarity function consistent with the SemVAE training metric yields optimal results. Furthermore, it demonstrates the complementary nature of the two terms: MSE ensures distribution-level precision, whereas cosine similarity enhances directional alignment, leading to better semantic matching and visual realism.

It is worth noting that the optimal settings identified in this ablation study differ slightly from the final hyperparameters presented in Table \ref{tab:hyperparam_settings}. This discrepancy arises because the ablation experiments were evaluated at 400K iterations; however, over the full training duration (4M iterations), the configuration detailed in Table \ref{tab:hyperparam_settings} yielded superior performance. Consequently, our final model adopts the settings from Table \ref{tab:hyperparam_settings} rather than strictly following the ablation outcomes.

\begin{figure*}[t]
  \centering
  \includegraphics[width=\textwidth]{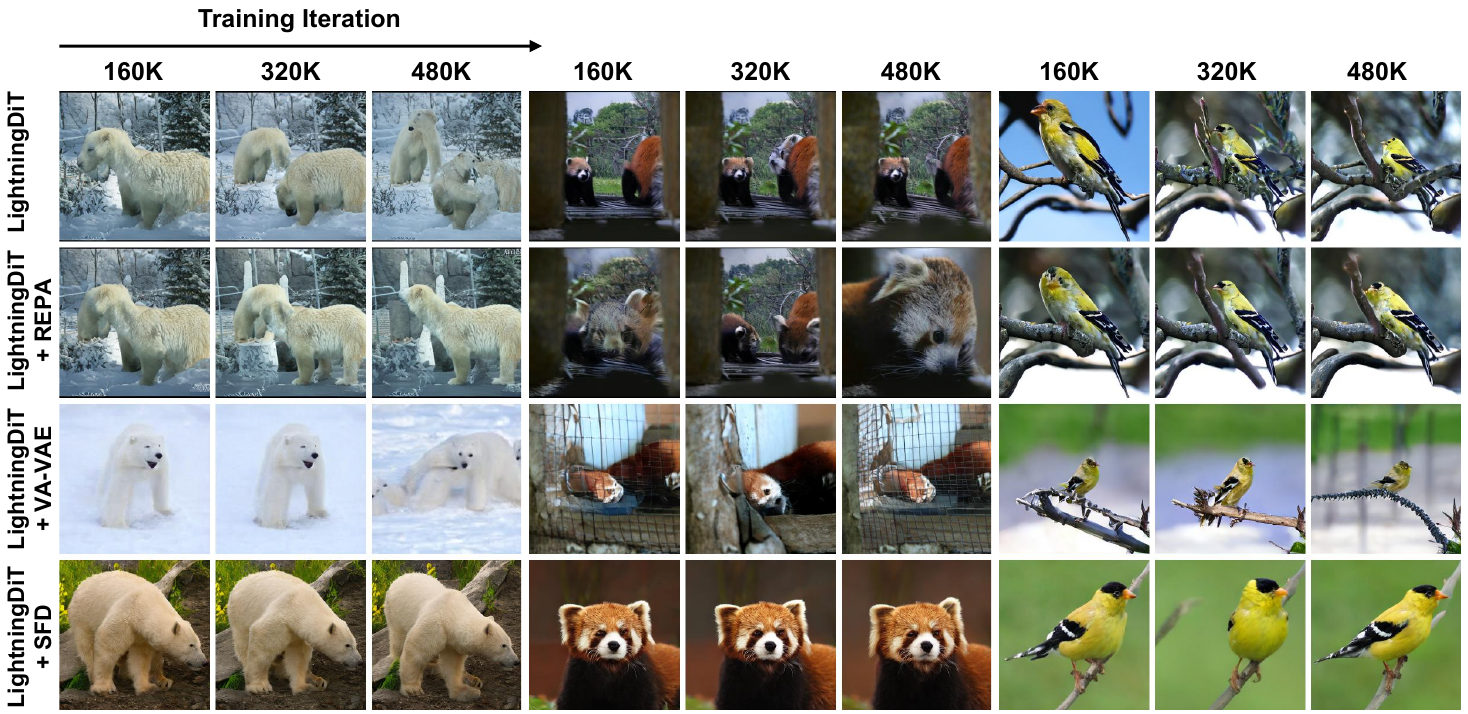}
  \caption{\textbf{Visualization of training results across different iterations (160K, 320K, and 480K).} Under a fixed random seed and identical initial noise, SFD produces clearer structures and more realistic details at early stages, demonstrating faster convergence compared with other variants.}
  \label{fig:converge_visual}
\end{figure*}

\section{Limitation and Future Work}
Currently, SFD employs a fixed temporal offset $\Delta t$ to manage the asynchronous denoising process. However, a static offset may not be optimal across all noisy levels. Future work could explore dynamic or adaptive schedules for $\Delta t$ to further enhance the synergy between semantic and texture generation. Furthermore, our framework presently relies on the REPA loss as an auxiliary objective to enforce feature alignment. A promising direction for future research is to investigate methods that eliminate the need for such auxiliary supervision, aiming for a cleaner and more streamlined optimization structure.

Beyond algorithmic refinements, extending and scaling SFD to more complex application scenarios represents a highly valuable research direction. Specifically, adapting SFD to text-to-image and text-to-video generation tasks could further validate its potential in handling intricate multimodal guidance and temporal consistency.

\section{More Visualization Results}
We qualitatively compare the training progression in Figure~\ref{fig:converge_visual}, where all models are evaluated using the same initial noise. The baseline LightningDiT, REPA, and VA-VAE variants exhibit weaker structural consistency and struggle to form coherent details in the early training stages. In contrast, SFD produces clearer structures and more realistic details at much earlier iterations, demonstrating noticeably faster convergence.

\begin{figure*}
  \centering
  \includegraphics[width=\textwidth]{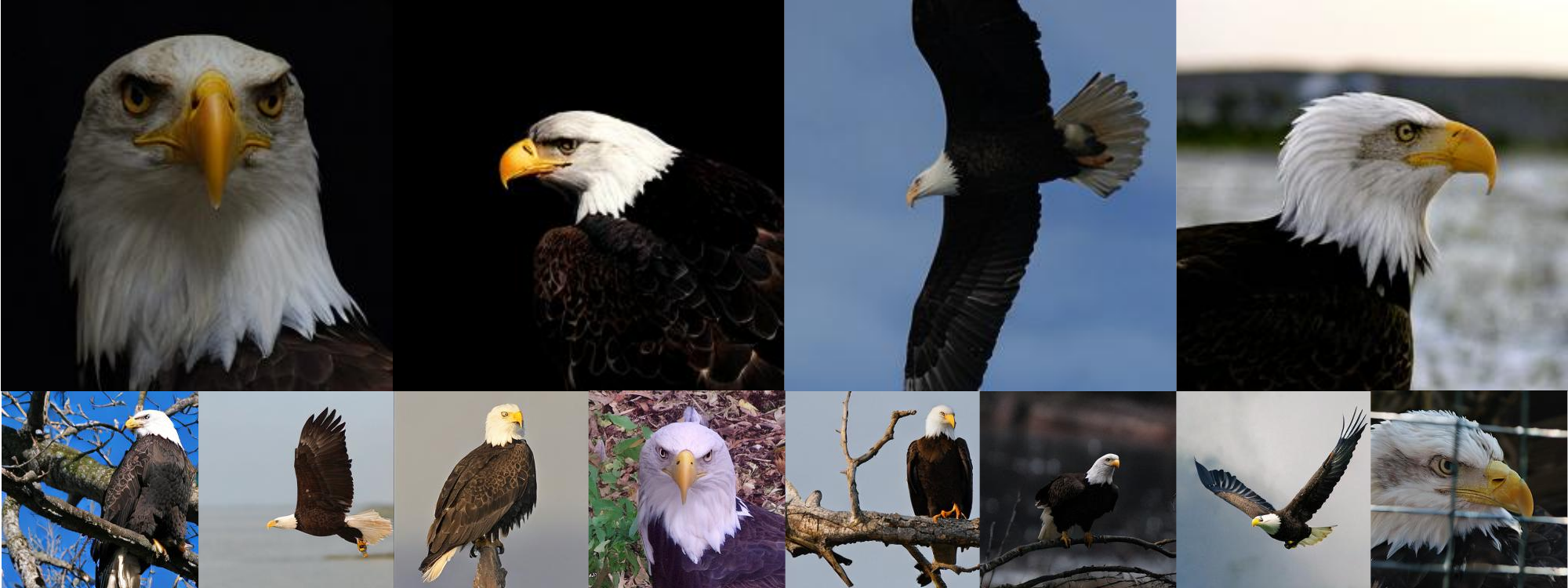}
  \caption{Visualization results of LightningDiT-XL + SFD for the ImageNet class ``Bald eagle'' (22).}
  \label{fig:demo_samples1}
\end{figure*}

\begin{figure*}
  \centering
  \includegraphics[width=\textwidth]{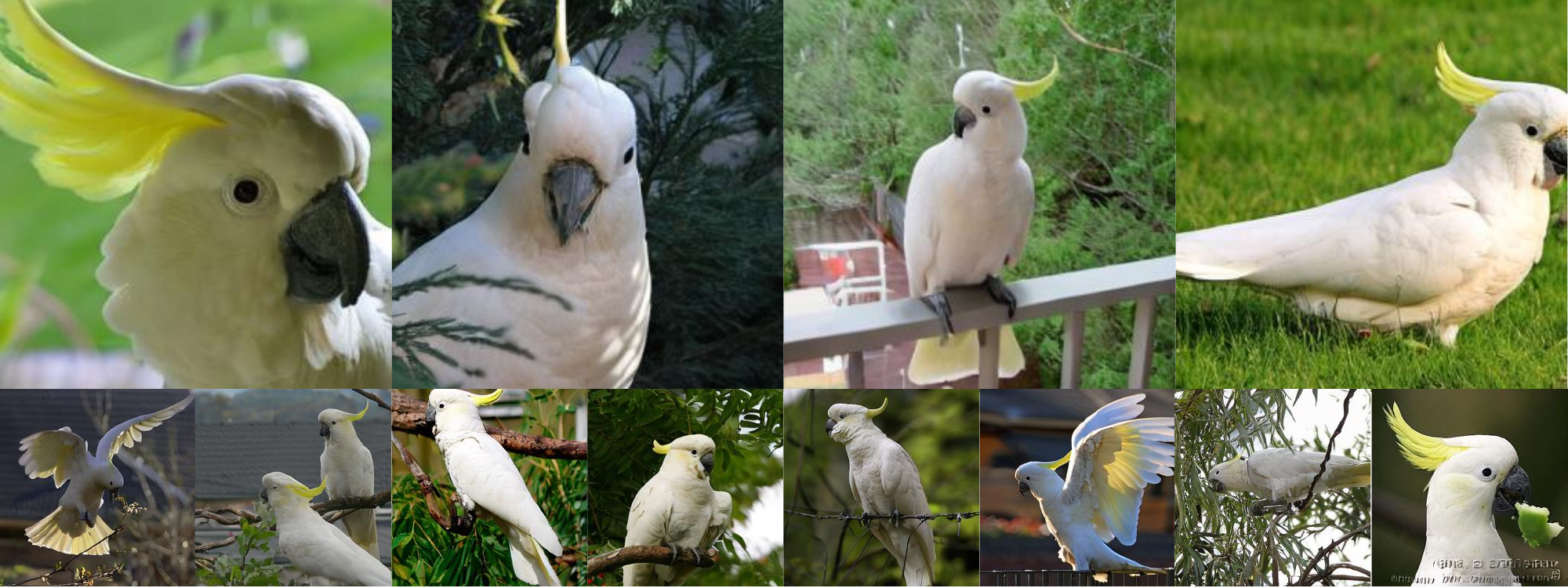}
  \caption{Visualization results of LightningDiT-XL + SFD for the ImageNet class ``Sulphur-crested cockatoo'' (89).}
  \label{fig:demo_samples2}
\end{figure*}

\begin{figure*}
  \centering
  \includegraphics[width=\textwidth]{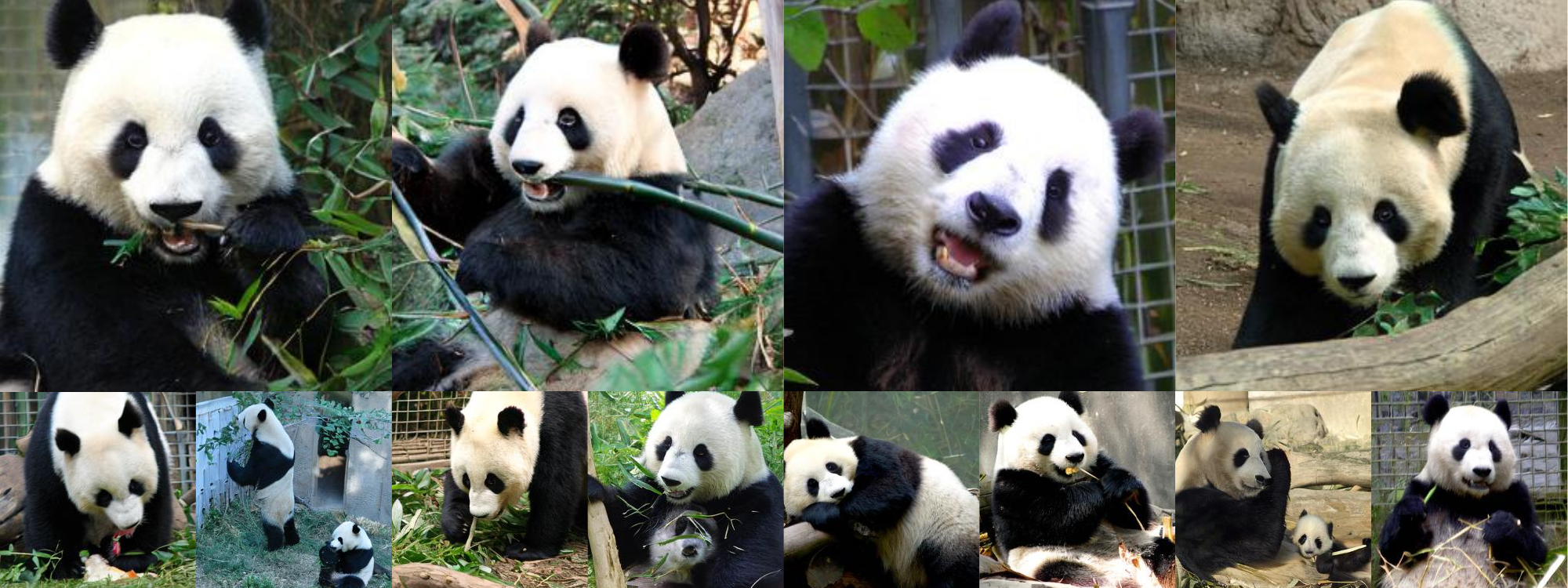}
  \caption{Visualization results of LightningDiT-XL + SFD for the ImageNet class ``Giant panda'' (388).}
  \label{fig:demo_samples3}
\end{figure*}

\begin{figure*}
  \centering
  \includegraphics[width=\textwidth]{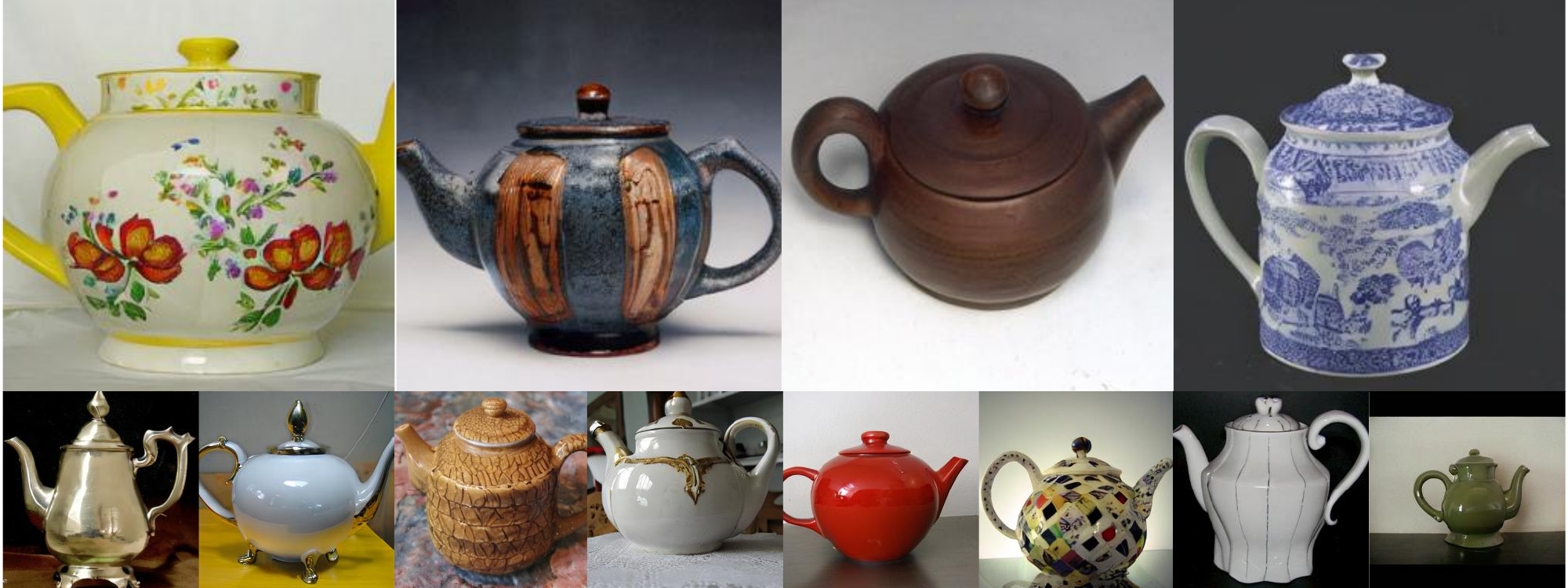}
  \caption{Visualization results of LightningDiT-XL + SFD for the ImageNet class ``Teapot'' (848).}
  \label{fig:demo_samples4}
\end{figure*}

\begin{figure*}
  \centering
  \includegraphics[width=\textwidth]{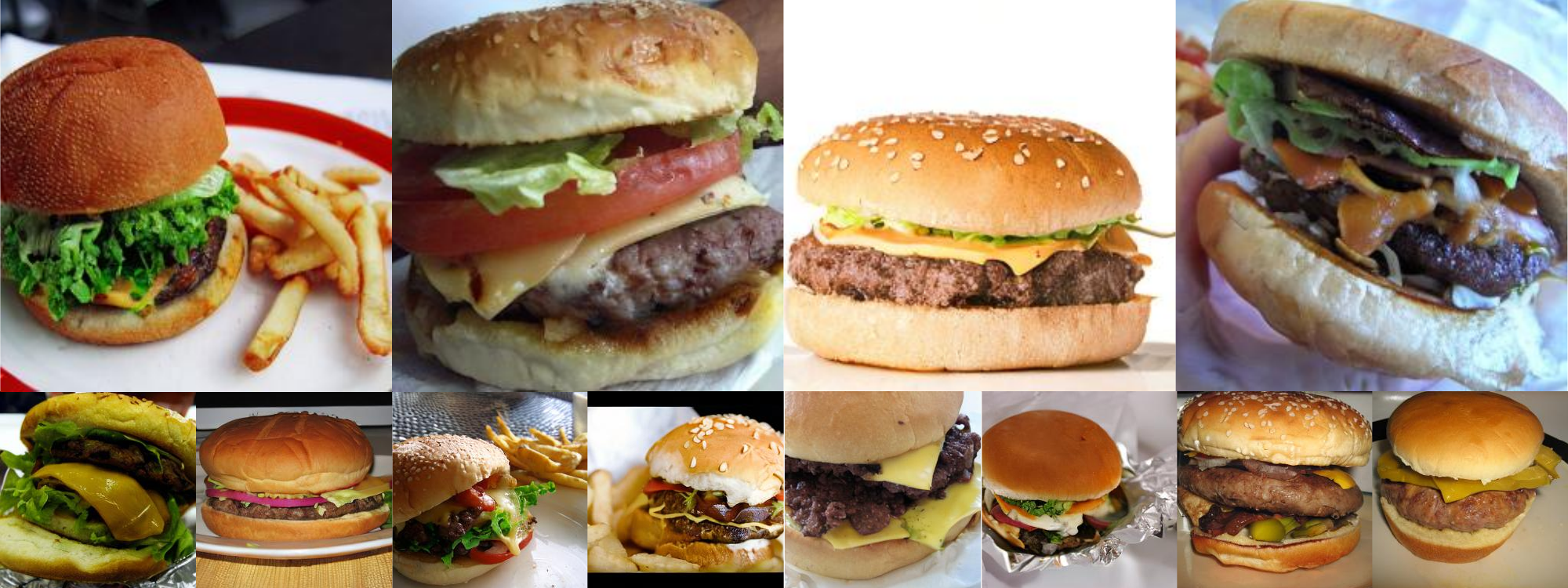}
  \caption{Visualization results of LightningDiT-XL + SFD for the ImageNet class ``Hamburger'' (933).}
  \label{fig:demo_samples5}
\end{figure*}

\begin{figure*}
  \centering
  \includegraphics[width=\textwidth]{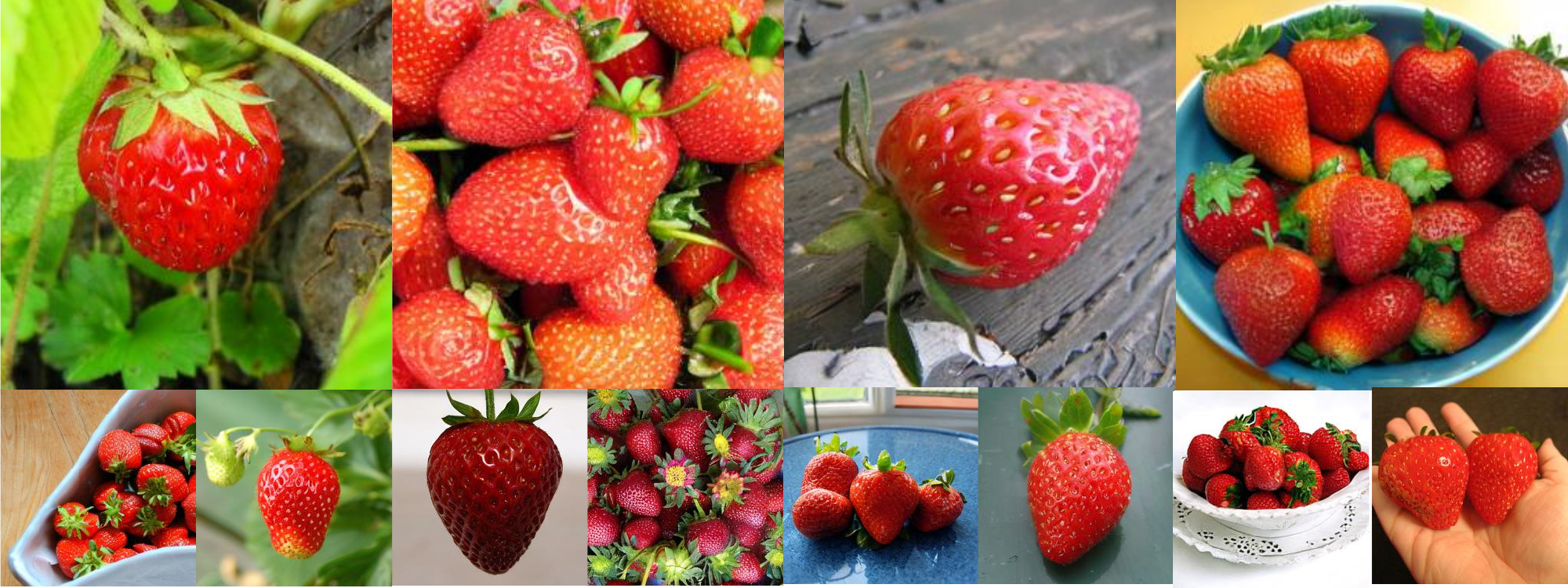}
  \caption{Visualization results of LightningDiT-XL + SFD for the ImageNet class ``Strawberry'' (949).}
  \label{fig:demo_samples6}
\end{figure*}

\begin{figure*}
  \centering
  \includegraphics[width=\textwidth]{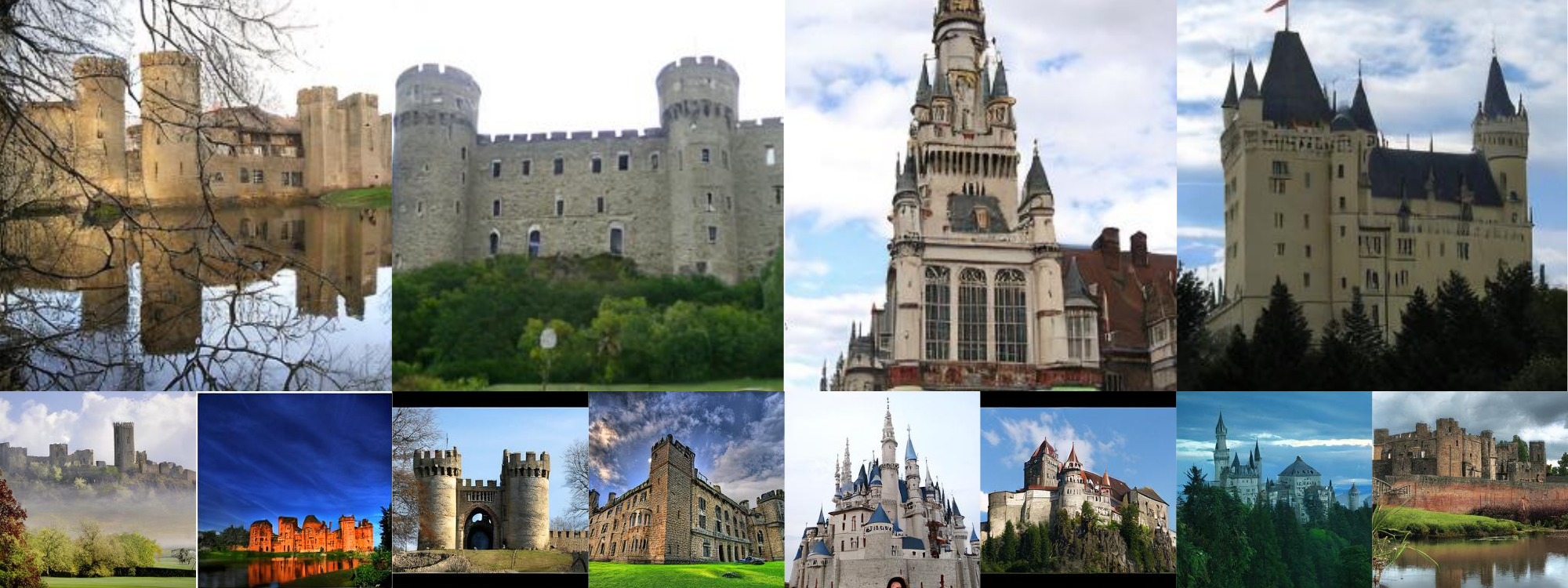}
  \caption{Visualization results of LightningDiT-XL + SFD for the ImageNet class ``Castle'' (483).}
  \label{fig:demo_samples7}
\end{figure*}

\begin{figure*}
  \centering
  \includegraphics[width=\textwidth]{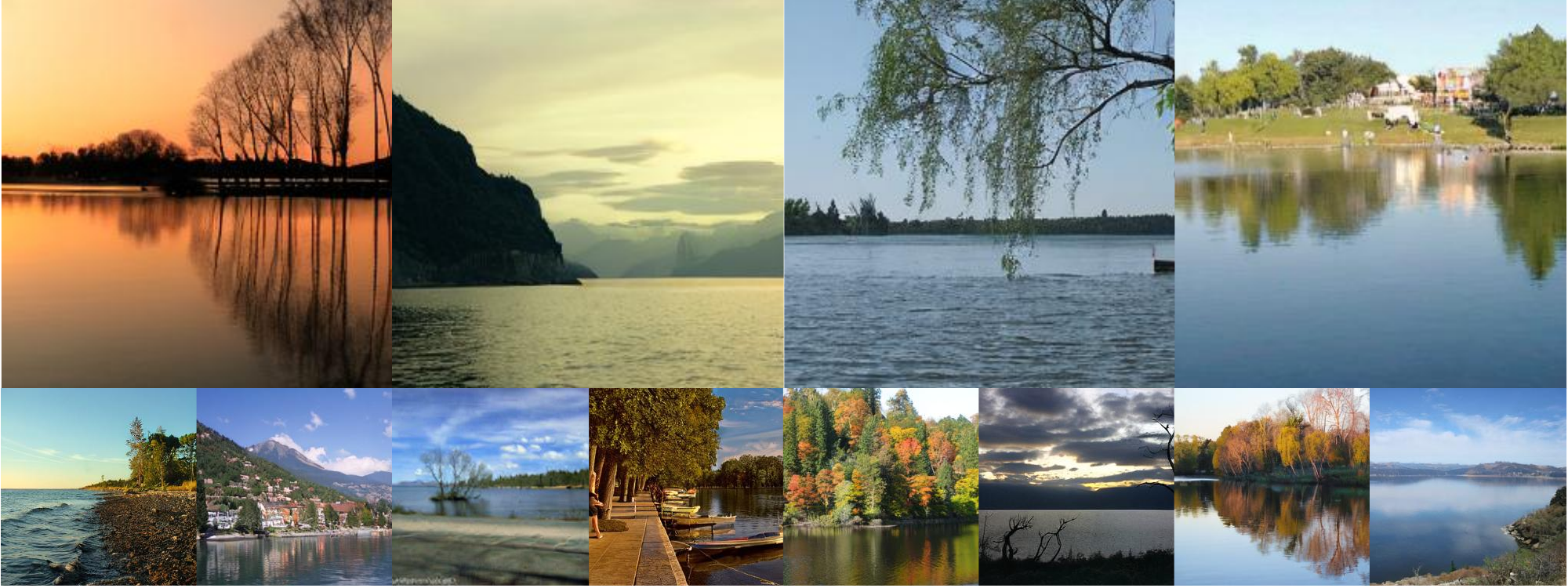}
  \caption{Visualization results of LightningDiT-XL + SFD for the ImageNet class ``Lakeside'' (975).}
  \label{fig:demo_samples8}
\end{figure*}

\begin{figure*}
  \centering
  \includegraphics[width=\textwidth]{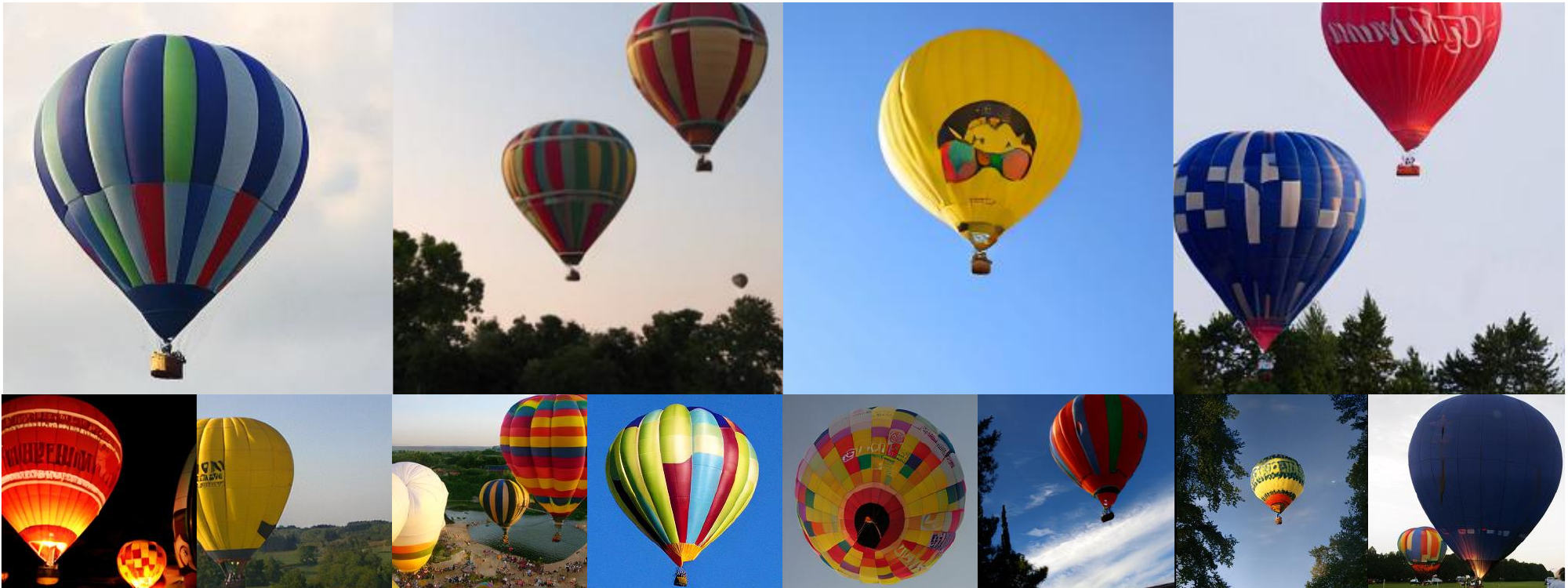}
  \caption{Visualization results of LightningDiT-XL + SFD for the ImageNet class ``Hot-air balloon'' (417).}
  \label{fig:demo_samples9}
\end{figure*}

We also present more visualization results of SFD in Figures~\ref{fig:demo_samples1} - \ref{fig:demo_samples9}.

